\PassOptionsToPackage{maths,theorems}{jmlrutils}
\documentclass[final,12pt]{clear2025} 

\title[Compositional Models for Estimating Causal Effects]{Compositional Models for Estimating Causal Effects}
\usepackage{times}

\newtheorem{assumption}{Assumption}

\usepackage{graphicx}
\usepackage{subcaption}  

\usepackage[english]{babel}

\usepackage[utf8]{inputenc} 
\usepackage[T1]{fontenc}    
\usepackage{hyperref}       
\usepackage{url}            
\usepackage{booktabs}       
\usepackage{amsfonts}       
\usepackage{nicefrac}       
\usepackage{microtype}      
\usepackage{xcolor}         
\usepackage{graphicx}
\usepackage{algorithm}
\usepackage{algorithmic}
\usepackage{subcaption}

\usepackage[english]{babel}

\clearauthor{\Name{Purva Pruthi} \Email{ppruthi@cs.umass.edu}\\
 \Name{David Jensen} \Email{jensen@cs.umass.edu}\\
 \addr College of Information and Computer Sciences \\ 
 University of Massachusetts Amherst \\
 Amherst, MA, USA}

\begin{document}

\maketitle
\begin{abstract}
Many real-world systems can be usefully represented as sets of interacting components. Examples include computational systems, such as query processors and compilers; natural systems, such as cells and ecosystems; and social systems, such as families and organizations. However, current approaches to estimating \textit{potential outcomes} and \textit{causal effects} typically treat such systems as single units, represent them with a fixed set of variables, and assume a homogeneous data-generating process. In this work, we study a \textit{compositional} approach for estimating individual-level potential outcomes and causal effects in structured systems, where each unit is represented by an \textit{instance-specific} composition of multiple heterogeneous components. The compositional approach decomposes unit-level causal queries into more fine-grained queries, explicitly modeling how unit-level interventions affect component-level outcomes to generate a unit’s outcome. We demonstrate this approach using modular neural network architectures and show that it provides benefits for causal effect estimation from observational data, such as accurate causal effect estimation for structured units, increased sample efficiency, improved overlap between treatment and control groups, and compositional generalization to units with unseen combinations of components. Remarkably, our results show that compositional modeling can improve the accuracy of causal estimation even when component-level outcomes are unobserved. We also create and use a set of real-world evaluation environments for the empirical evaluation of compositional approaches for causal effect estimation and demonstrate the role of composition structure, varying amounts of component-level data access, and component heterogeneity in the performance of compositional models as compared to the non-compositional approaches.
\end{abstract}

\section{Introduction}
Many applications require estimating individual-level potential outcomes and treatment effects, including personalized medicine \citep{curth2024using}, individualized instruction \citep{kochmar2022automated}, and custom online advertising \citep{bottou2013counterfactual}. Standard approaches to heterogeneous treatment effect estimation \citep[e.g., ][]{hill2011bayesian, athey2016recursive, wager2018estimation, chernozhukov2018double} typically assume that the units of analysis can be represented by a fixed set of random variables that are sampled from a fixed causal graph, following a homogeneous data generating process, known as unit homogeneity assumption \citep{holland1986statistics}. 

However, many real-world systems are \textit{heterogeneous} and \textit{modular} --- they decompose into heterogeneous functional \textit{components} that interact in various ways to produce system behavior \citep{callebaut2005modularity, johansson2022generalization}. Input data to such systems is often structured, variable-sized, and sampled from different causal graphs,
making it challenging to reason about the system’s behavior. An alternative approach to analyzing such systems is to exploit their \textit{compositionality}, assuming that the system behavior can be understood in terms of the behavior of familiar \textit{re-usable} components
and how they are composed. Estimating individual treatment effects for compositional systems is an important problem, particularly as the complexity of modern technological systems increases. Modern computational systems such as databases, compilers, and multi-agent systems can generate large amounts of experimental and observational data containing fine-grained information about the structure and behavior of modular systems, which often remains unused by the existing approaches for estimating causal effects.

\begin{figure}[h]
    \centering
\includegraphics[width=\linewidth]{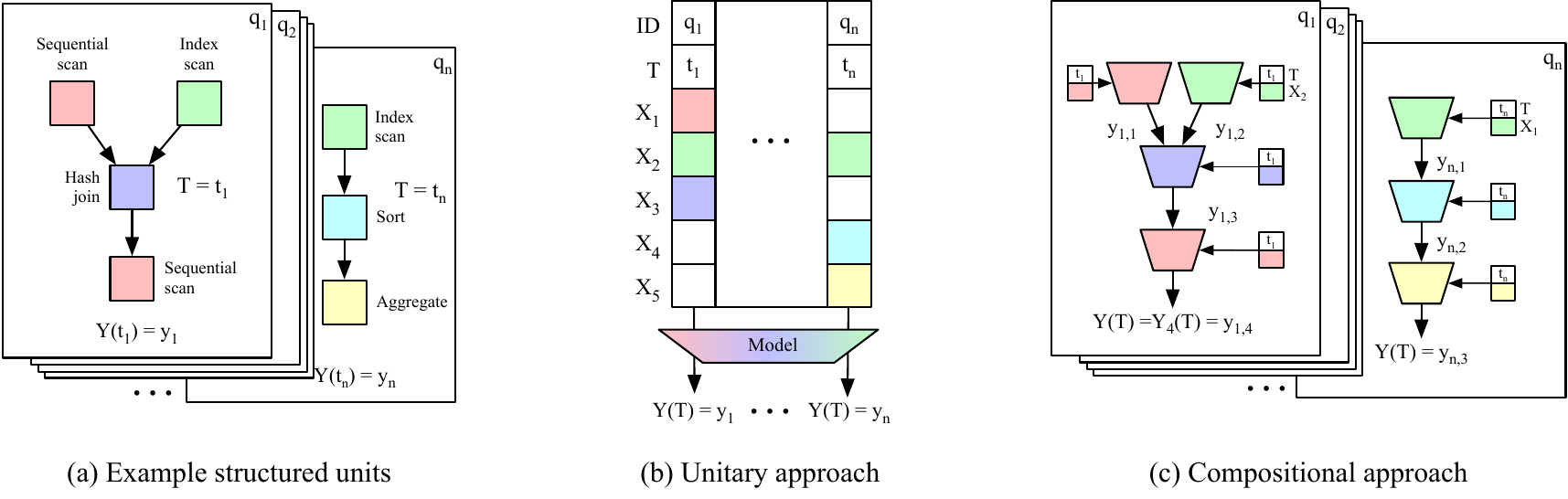}
    \caption{\textbf{Overview of key ideas:} (a) \textbf{\textit{Structured units:}} Units are composed of multiple heterogeneous components. Each color represents a distinct component. Treatment $T$ is applied to the unit, and the compositional system processes the inputs under intervention, returning potential outcomes. (b) \textbf{\textit{Unitary approach:}} Standard approaches to effect estimation flattens the underlying structure. They use a fixed-size representation for each unit, aggregating component-level information to estimate unit-level potential outcomes. (c) \textbf{\textit{Compositional approach:}} The compositional approach models each unit with an instance-specific structure. Component-level covariates $X_j$ and outcomes $Y_j$ are used to train each component model, and component-level outcomes are hierarchically aggregated to estimate unit-level potential outcomes. Each color represents a distinct component model with different parameters.}
    \label{fig:intro_graphics}
\end{figure}

Figure \ref{fig:intro_graphics} provides a schematic overview of causal inference for compositional systems and how it is addressed by different approaches to causal estimation. Consider a relational database query execution system with component operations such as \texttt{scan}, \texttt{sort}, \texttt{aggregate}, and \texttt{join}. The system takes input including tables (e.g., A, B) and a query execution plan (e.g., \texttt{scan(join(scan(A),scan(B)))}) and returns a new table as output. Query executions represent the unit of analysis, and query plans explicitly describe the \textit{compositional} and \textit{hierarchical} structure of those units --- query executions consist of heterogeneous component operations that can be combined in a vast number of different ways. In addition, the compositional structure of query execution is \textit{instance-specific} --- the number, kind, and structure among components may differ across each unit. These units can be represented as hierarchical graphs (e.g., parse-trees), where each node is a component operation and edges represent the information flow between the components. 

Given such a system, consider modeling the causal effect of memory size on execution time for different query plans. This problem can be formulated as using observational data to estimate the \textit{individual-level} effects of interventions on structured units.\footnote{Here, individual-level effect estimation refers to conditional average treatment effect (CATE) estimation and heterogeneous treatment effect estimation \citep{athey2016recursive, shalit2017estimating, kunzel2019metalearners}.} In real-world data on query execution, interventions such as memory size might be chosen based on the structure and features of the query, making the data \textit{observational}. In the terminology of causal inference, each query execution is a \textit{unit of analysis}, the features of the relations are \textit{pre-treatment covariates}, memory size is the \textit{intervention}, and execution time is the \textit{potential outcome} \citep{rubin1974estimating, rubin2005causal}. We might also want to predict the effects for a population of arbitrary query executions that contain novel combinations of the component operations. In that case, it is desirable for the learned models to \textit{compositionally generalize} to units with unseen combinations of the components. Other real-world use-cases of causal reasoning in structured data are discussed in the supplementary material (Section \ref{sec:real_world_examples}).

Standard approaches to heterogeneous treatment effect estimation \citep[e.g., ][]{hill2011bayesian, athey2016recursive} typically ignore the underlying structure and represent each compositional unit using a fixed-size feature representation, which poses several estimation and identifiability challenges. As the structure and complexity of each unit vary, estimating effects at the unit level requires reasoning about the similarity among the heterogeneous units in a high-dimensional space. Additionally, representing all the units with the same features leads to sparse feature representation and aggregation of the features of multiple instances of each component, causing identifiability challenges. We use the term \textit{unitary models} to denote these approaches that exclusively model unit-level quantities.

In contrast, we study a \textit{compositional} approach to causal effect estimation for structured units from observational data. This approach estimates the component-wise potential outcomes using the observational data available for that component, pooled across each instance of the component among units. It then forms an estimate of the unit-level potential outcomes and treatment effects by aggregating the component-level estimates according to the given compositional structure.
This approach facilitates construction of \textit{instance-specific} causal models (models whose structure changes based on the specific components present in the specific units being modeled) using {\textit{modular} neural network architectures} to explicitly represent the components of each unit of analysis (Figure \ref{fig:intro_graphics}(c)).\footnote{Modular neural network architectures are chosen as an implementation choice to demonstrate a compositional approach, but component-wise causal estimation can be done using a variety of parametric and non-parametric model classes.} \textbf{We formalize a novel \textit{compositional} framework in the context of causal effects and potential outcomes} to facilitate the study of the compositional approach and provide a detailed analysis of the unique benefits and costs of such an approach for accurate unit-level CATE estimation. We focus on the case of hierarchically structured compositional systems without feedback and with simple interactions because this represents the minimal compositional system necessary to understand the key characteristics of the compositional approach and compare it to unitary modeling approaches.  

We show that the compositional approach provides several novel benefits for causal inference from observational data. The instance-specific model allows \textit{scalable} causal effect estimation for variable-size units by greatly reducing the inherent dimensionality of the task. Instance-specific modular architectures are widely used in associational machine learning for modeling large-scale natural language data, structured vision data, and sequential decision-making, providing sample efficiency and computation benefits \citep{shazeer2016outrageously, pfeiffer2023modular}. However, only a relatively sparse body of work in causal inference has focused on using instance-specific modular models using hierarchical and relational data \citep{maier2013sound, lee2016learning, salimi2020causal,ahsan2023learning}, and even this work has been severely hampered by the lack of available data from compositional domains. To address this gap, \textbf{we introduce three novel and realistic evaluation environments to evaluate compositional approaches for causal effect estimation} --- query execution in relational databases, matrix processing on different types of computer hardware, and simulated manufacturing assembly line data based on a realistic simulator. 

The modular structure incorporated in the compositional approach facilitates effect estimation for units with unseen combinations of components, enabling \textit{compositional generalization}. In various fields of machine learning --- computer vision \citep{andreas2016neural}, language \citep{hupkes2020compositionality}, reinforcement learning \citep{peng2019mcp}, and program synthesis \citep{shi2024exedec}, researchers have studied the compositional generalization capabilities of the modular approaches as compared to non-modular approaches for prediction tasks \citep{bahdanausystematic, jarvisspecialization}. However, a study of the benefits of the compositional approach compared to the standard approaches is missing in causal inference. \textbf{We study the relative \textit{compositional generalization} capabilities of compositional and unitary approaches in estimating individual treatment effects for novel units.}    

Individual effect estimation from observational data requires assumptions such as \textit{ignorability} and \textit{overlap} \citep{pearl2009causality, rubin2005causal}. Satisfying the overlap assumption becomes challenging as the dimensionality of covariates increases \citep{d2021overlap}. Learning lower-dimensional representations of data that satisfies the ignorability and overlap assumption is desirable in such situations \citep{johansson2016learning, johansson2022generalization}. Exploiting the compositionality of the underlying data-generating process is one way to learn a lower dimensional representation, allowing better overlap between treatment groups. \textbf{We show that the compositional approach performs better than the unitary approach as the distributional mismatch between the treatment and control groups increases, especially in cases where treatment is assigned based on the unit's structure.}

Despite these potential benefits, learning compositional models for effect estimation has pitfalls, including larger numbers of parameters to estimate, sensitivity to individual components, and errors in modeling component interactions. In this paper, \textbf{we analyze the role of component-level data access, composition structure, and heterogeneity in component function complexity in the relative performance of the compositional approach}. 
For example, we observe that compositional and unitary approaches perform similarly when modeling systems with homogeneous component functions---such as matrix processing---where a single component (e.g., matrix multiplication) dominates the overall unit-level outcome and all component outcome functions belong to the same polynomial class. See Section \ref{sec:findings} for additional pitfalls.

\textbf{Note:} We include a detailed discussion of the related work in the appendix (Section \ref{sec:related work}).

\section{Compositional Framework for Causal Effect Estimation}
\label{sec:compositional_framework}
Below, we describe the compositional data-generating process in modular systems, provide the estimands of interest at the unit and component levels, and discuss identifiability assumptions.

\textbf{Preliminaries:} Assume that each unit $i$ has pre-treatment covariates $\mathbf{X}_i = \mathbf{x} \in \mathcal{X} \subset \mathbb{R}^d$, a binary treatment $T_i \in \{0,1\}$, and two potential outcomes $\{Y_i(0), Y_i(1)\}\in \mathcal{Y} \subset \mathbb{R}$ \citep{rubin1974estimating, rubin2005causal}. In observational data, we only observe one of the potential outcomes for each unit, $Y_i  = Y_i(T_i)$, known as the \textit{observed} or \textit{factual} outcome. The missing outcomes ${Y_i}_{CF} = Y_i(1-T_i)$ are known as \textit{unobserved} or \textit{counterfactual} outcomes. Conditional average treatment effect (CATE) is defined as $\tau(x): \mathbb{E}[Y_i(1) - Y_i(0) | \mathbf{X}_i = \mathbf{x}]$. Estimating CATE requires assumptions of ignorability, overlap, and consistency \citep{rosenbaum1983central}. Under these assumptions, $\tau(x)$ is identifiable by $\tau(\mathbf{x}) = \mathbb{E}[{Y_i|\mathbf{X}_i = \mathbf{x}, T=1}] - \mathbb{E}[{Y_i|\mathbf{X}_i=\mathbf{x}, T=0}]$ \citep{pearl2009causality}. CATE estimation typically uses direct outcome modeling with treatment as a feature, separate regression models \citep{kunzel2019metalearners}, or propensity score methods \citep{kennedy2023towards}. We illustrate the compositional approach by directly estimating the potential outcomes using shared treatment. 


\subsection{Compositional data generating process} 
Consider a compositional system with $k$ distinct and heterogeneous classes of components: $\mathcal{M} = \{M_1, M_2, \dots M_k\}$. All units share this set of reusable components. Each structured unit $Q_i: (G_i, \{\mathbf{X}_{ij}\}_{j=1:m_i})$ is described using an interaction graph $G_i$ and a set of component-specific covariates $\{\mathbf{X}_{ij}\}_{j=1:m_i}$, where $m_i$ denotes the number of components in unit $i$.  Note that the number of components $m_i$ can be greater than the number of distinct components $k$ in the system, indicating the presence of multiple instances of each component class in some or all units. 

The graph $G_i = (C_i, E_i)$ is a directed hierarchical tree representing component interactions (Figure \ref{fig:intro_graphics}(a)), with nodes $C_i$ and edges $E_i$. Each unit $i$ contains components $C_i = \{c_1, c_2, \dots c_{m_i}\}$, where each component belongs to a class $c \in M_o, o \in \{1, 2, \dots k\}$. $G_i$ defines the processing order of $m_i$ components for unit \textit{i}, with instance-specific structure varying across units. Components process structured units from top-to-bottom, with final output from the bottom-most node. For an edge $c_j \rightarrow c_{j'}$, $c_j$ is the parent and $c_{j'}$ the child component. $Pa_{G_i}(c_{j'})$ denotes indices of components with direct edges to $c_{j'}$. Component $j$'s covariates are $\mathbf{X}_{ij} \in \mathbb{R}^{d_j}$, where subscript $i$ denotes the unit and $j$ denotes the component instance.

\textbf{Shared treatment and treatment assignment mechanism in structured units:} A unit-level treatment $T_i$ is selected for each unit, affecting the potential outcomes of some or all components through shared or distinct mechanisms. While the compositional framework allows component-specific treatment analysis, we focus on unit-level treatments to facilitate a direct comparison of the compositional and unitary approaches. The treatment assignment mechanism $P(T_i=1|Q_i=q)$ depends on both the graph structure and joint covariate distribution, introducing two sources of observational bias: (1) distribution shift among the covariates; and (2) distribution shift in the structure and composition of the components between treatment groups. 

\textbf{Unit-level outcome and fine-grained outcomes:} 
Let $Y_{i}(t)$ denote the unit-level and $\{Y_{ij}(t)\}_{j=1:m_i}$ denote the component-level potential outcomes under treatment $t$ for unit $Q_i$. The interaction graph $G_i$ also defines the causal dependencies among potential outcomes.  We make the causal Markov assumption for components in the graph $G_i$ that the potential outcome of a component $j$ directly depends on the component's covariates $\textbf{X}_j$, shared treatment $T_i$ and the outcomes from the parent components. 
For component class $o \in \{1, 2, \dots k\}$, let $\mu_{ot}$ denote the ground-truth \textit{expected} potential outcome function, shared across instances of the same component.  If component-level noise is assumed to be zero-mean random variables $\epsilon_o(0), \epsilon_o(1)$ and $c_j \in M_o$,\footnote{Additive noise simplifies modeling conditional distributions of component potential outcomes in Section \ref{section:hierarchical_composition_dist}, though the compositional approach also extends to non-additive noise.} the data-generating process for $Y_{ij}(t)$, $j \in \{1, 2, \dots m_i\}$ and $t \in \{0,1\}$:

\begin{equation}
    Y_{ij}(t) = \mu_{ot}(\mathbf{X}_{ij}, \{Y_{il}(t)\}_{\{l \in {Pa(c_j)}\}}) + \epsilon_{io}(t). 
\label{eqn:Markov_dependence}
\end{equation}

The unit-level potential outcome is generated by aggregating component outcomes via an instance-specific function $g$: $Y_i(t) = g(Y_{i1}(t), Y_{i2}(t) \dots Y_{im_i}(t), G_i)$.
For composition in the hierarchical graph, the data-generating process of each component's outcome already takes the input from the parent's outcome, following Markov dependence (Equation \ref{eqn:Markov_dependence}). We can define these outcomes cumulatively, meaning we aggregate them up as we go along, so that the final component's outcome represents the complete unit-level outcome, $Y_i(t) = Y_{im_i}(t)$, where $m_i$ indicates the last component in $G_i$. For example, consider query execution time: rather than measuring each component's time separately, we can measure the accumulated time of each component and all its parent components. In Figure \ref{fig:intro_graphics}(c), this cumulative approach means we can use the final sequential scan's outcome for $Q_1$ as the total execution time of the unit. This formulation allows us to learn the instance-specific aggregation functions and unit outcomes as part of learning the component potential outcomes.

\textbf{Note on a formal definition of causal compositionality:} A mathematically precise definition of compositionality is an active research area in machine learning \citep{ram2024makes, elmoznino2025towards}. Given the state of the current literature, our work adopts a data-generating process view of compositionality inspired by real-world computational systems with explicit causal mechanisms. More specifically, compositionality in our work is defined through: (1) structured units with instance-specific compositions; and (2) component interactions that are formalized through an interaction graph and the Markov assumption (Equation \ref{eqn:Markov_dependence}). Previous work in machine learning has used a similar data-generating process view to develop compositional approaches \citep{andreas2016neural, wiedemer2024compositional}. We also provide a graphical plate notation-based representation of the compositional data generating process in Section \ref{sec:plate_based_notation}.

\subsection{Unitary representation of compositional data}

As already mentioned,  an alternative to a compositional model is a unitary model in which units are represented by a single, fixed-size, high-dimensional feature vector, $\mathbf{X}_i \in \mathbb{R}^d$ that represents some aggregation of the component level input features $\{\mathbf{X}_{ij}\}_{j=1}^{m_i}$. For simplicity in our experiments, we assume the mean aggregation function (i.e., for $N_o$ occurrences of component class $o$ in unit $i$, $\mathbf{X}_{io} = \frac{1}{N_o}\sum_{j, c_j \in M_o} \mathbf{X}_{ij}$). Similarly, corresponding outcomes are also aggregated. To fairly compare the unitary and compositional approaches, we also incorporate structural information by including the number of instances of each component $N_{jl}$ present in each unit at each tree depth with maximum depth $L$: $\mathbf{X} = [\mathbf{X}_1, \mathbf{X}_2, \dots \mathbf{X}_k, N_{11}, N_{12} \dots N_{kL}]$.

\subsection{Causal Estimands}
The CATE for a structured unit $q$ is conditional on both the structure $G_i$ and the set of
component-level features $\{\mathbf{x}_{ij}\}_{j = 1:m_i}$.  Taking the conditional expectation with respect to the structure and variable-length representation of the units allows a more accurate definition of the CATE for compositional units than the standard unitary representation. For ease of notation to describe conditional distributions, we denote the combined inputs to a component $j$ as $\mathbf{Z}_j(t) = (\mathbf{X}_j, \{Y_l(t)\}_{l \in {Pa(C_j)}})$.
\begin{definition}
    The conditional-average treatment effect (CATE) estimand for structured input $Q_i=q$ is defined as: $\tau(q) = \mathbb{E}[Y_i(1) - Y_i(0) | Q_i = q] = \mathbb{E}[Y_i(1) - Y_i(0) | Q_i = (G_i, \{\mathbf{x}_{ij}\}_{j = 1:m_i})]$
\end{definition}

\begin{definition}
    The conditional-average treatment effect (CATE) estimand for component $j$ with $\mathbf{X}_j = \mathbf{x}_j \in \mathbb{R}^{d_j}$ is defined as: $\tau(\mathbf{z}_j) = \mathbb{E}[Y_j(1) - Y_j(0) | \mathbf{Z}_j = \mathbf{z}_j]$.
\end{definition}
We define the component-wise  distributions as $P(Y_j(t)|\mathbf{Z}_j(t) = \mathbf{z}_j)$. In hierarchical composition, the unit outcome equals the final component outcome: $\mathbb{E}[Y_i(t) |Q_i = q] = \mathbb{E}[Y_{im_i}(t)|Q_i=q]$. This conditional expectation can be expressed by marginalizing intermediate component outcomes using the Markov assumption (Equation \ref{eqn:Markov_dependence}). 

$\mathbb{E}[Y_i(t) |Q_i = q] = \int_{Y_{im_i-1}(t)} \int_{Y_{im_i-2}(t)} \dots \int_{Y_{i1}(t)}\mathbb{E}[Y_{im_i}(t)|\mathbf{Z}_{im_i}(t)] \prod_{j=1}^{m_i-1} P(Y_{ij}(t)|\mathbf{Z}_{ij}(t))$ 

We use the following nested expectation expression as shorthand for marginalization over intermediate component outcomes: $\mathbb{E}[Y_i(t) |Q_i = q] = \mathbb{E}_{Y_{i1:im_i-1}(t)}[\mathbb{E}[Y_{im_i}(t)|\mathbf{Z}_{im_i}(t)]]$.

\subsection{Identifiability assumptions}
The key identifiability assumptions for component-level causal estimands are similar to those for unit-level estimands --- ignorability, consistency, and overlap. However, in structured units having multiple heterogeneous components and instance-specific composition, it is more plausible for these assumptions to hold for the component level rather than the unit level, particularly for ignorability and overlap. 



\textbf{Ignorability:} Component-level ignorability assumes that component level potential and assigned treatment are independent conditioned on the components' covariates, i.e.,  $Y_{j}(1), Y_{j}(0) \perp T | \mathbf{X}_{j}$. Component-level ignorability is based on the intuition that components are distinct heterogeneous sub-systems that are specialized to process parts of the whole input. This suggests that a subset of the unit-level high-dimensional covariates is sufficient to predict a component's outcome, even when treatment might have been assigned based on the structure of the components or joint distribution of the component covariates. The component-level potential outcomes depend on both the component's pre-treatment covariates and the potential outcomes of the parent components. As the treatment is assigned before any component's potential outcomes are observed, we can assume that the component's covariates $\mathbf{X}_j$ are sufficient to satisfy ignorability assumptions.

\textbf{Overlap:} Component-level overlap assumes that overlap holds for the component level covariates $\mathbf{X}_j = \mathbf{x}_j$, i.e., $\forall \mathbf{x}_j \in \mathcal{X}_j, t \in \{0,1\}: \  0 < p(T=t|\mathbf{X}_j=\mathbf{x}_j) < 1$.  When unit-level overlap holds, component-level overlap is implied automatically for the feature subset. In compositional data, there can be two sources for distribution mismatch. (1) \textit{Structure-based treatment assignment:} Suppose treatment depends strongly on graph structure ($P(T=1|G = g_1) = 1$, $P(T=1|G = g_2) = 0$). For unitary representation including structural features $N_{jl}$, the overlap assumption is violated for units with the structures as $g_1$ and $g_2$, while overlap is maintained for compositional approaches as the structure is incorporated as part of the model rather than input representation. If we exclude structural information from unitary representation, then overlap is satisfied, but ignorability is violated because the structure is a confounder affecting both $T_i$ and $Y_i$. (2) \textit{Covariate-based treatment assignment:} When treatment depends on the covariate distribution, both approaches' identifiability relies on overlap quality. Due to the compositional nature of units, violation of overlap for a component's covariates violates overlap for the unit-level and vice versa. However, due to the lower dimensionality of the component's covariates, the degree of distribution mismatch between the covariates is likely to be lower than the unit-level high-dimensional covariates. 

\textbf{Identifiability for hierarchical composition:} The CATE estimand for a structured unit $Q_i=q$ is identified by the following: $\tau(q) = \mathbb{E}_{Y_{1:m_i-1}}[\mathbb{E}[Y_{m_i}|\mathbf{Z}_{m_i} = \mathbf{z}_{m_i}, T = 1]] - \mathbb{E}_{Y_{1:m_i-1}}[\mathbb{E}[Y_{m_i}|\mathbf{Z}_{m_i} = \mathbf{z}_{m_i}, T = 0]]$
if we assume Markov dependence assumption (Equation \ref{eqn:Markov_dependence}), component-wise ignorability, overlap, and consistency. The proof is provided in the supplementary material.

\textbf{Additive parallel composition (special case):} A special case of the interaction graph $G_i$ is when all the components are independent and parallely compute the potential outcomes. This condition can be expressed as when the potential outcomes of components are \textit{conditionally independent} given the component's covariates, i.e., $Y_j(t) \perp Y_l(t) | \mathbf{X}_j \forall j \in \{1, \dots k\}, j \neq l$.  Suppose the aggregation function is a linear combination of the component's outcomes, e.g., addition. The unit-level CATE can be expressed as the sum of the potential outcome estimands. 
$Y_{ij}(t) = \mu_{ot}(\mathbf{X}_{ij}) + \epsilon_{io}(t)$, $Y_{i}(t) = \sum_{j=1}^{m_i} Y_{ij}(t)$. In that case, the CATE can be directly expressed as the linear combination of the component-level CATE and is identified as follows: $\tau(q) = \sum_{j=1}^{m_i} \tau(\mathbf{x}_{ij}) = \sum_{j=1}^{m_i} \mathbb{E}[Y_{ij} |\mathbf{x}_{ij}, T_i = 1] - \mathbb{E}[Y_{ij} |\mathbf{x}_{ij}, T_i = 0]$. The proof is provided in the supplementary material. 
Parallel composition appears across machine learning domains---from independent composition of latent object attributes in computer vision \citep{higgins2018scan, wiedemer2024compositional} to spatial skill composition in reinforcement learning \citep{van2019composing}. Similarly, sequential composition appears in the options framework in reinforcement learning \citep{sutton1999between}. Our work examines these varied composition structures to understand their impact on compositional generalization.

\section{Learning compositional models from observational data}
Below, we discuss the algorithm for learning the hierarchical composition model from observational data, given different amounts of information about the component's covariates and outcomes. We include algorithms for parallel structured composition models in Section \ref{section:composition_models_app}. 

\subsection{Hierarchical Composition Model}
The below algorithm facilitates the modeling of noise variables  $\epsilon_{ot}$ (assuming zero-mean additive noise variables in Equation
1) along with the \textit{expected} potential outcome functions $\mu_{ot}$. This allows the marginalization of intermediate component-level potential outcomes to obtain unit-level CATE estimates for sequential and hierarchical compositions. 

\label{section:hierarchical_composition_dist}
\textbf{Model Training:} The component models for estimating mean and variance of conditional distribution of the component-level potential outcomes for component class $o \in \{1, 2, \dots k\}$ are denoted by $(\hat{f}_{\theta_o}, \hat{\sigma}^2_{\psi_o}): \mathbb{R}^{d_o} \times \mathbb{R}^{D} \times \{0,1\} \rightarrow \mathbb{R} \times \mathbb{R}^+$, assuming maximum in-degree of the graph $G_i$ as $D$. Each model corresponding to component class $o$ is parameterized by separate and independent parameters $\theta_o$ for the mean and $\psi_o$ for the variance.
For a given observational data set with $n$ samples, $\mathcal{D}_F = \{q_i, t_i, y_i\}_{i=1:n}$, we assume that we observe component-level features $\{\mathbf{x}_{ij}\}_{j=1:m_i}$, assigned treatment $t_i$ and fine-grained component-level potential outcomes $\{y_{ij}\}_{j=1:m_i}$ along with unit-level potential outcomes $y_i$. If component instance $c_j \in M_o$, training of each component model $o$ involves the independent learning of the parameters by minimizing the \textit{negative log-likelihood loss}:
\begin{equation}
    (\theta^{*}_o, \psi^{*}_o):= \arg\min{\theta_o, \psi_o} \frac{1}{N_o}\sum_{m=1}^{N_o} \left[\frac{1}{2}\log(2\pi\hat{\sigma}^2_{\psi_o}(\mathbf{z}_{m}, t_m)) + \frac{(y_{ij} - \hat{f}_{\theta_o}(\mathbf{z}_{m}, t_m))^2}{2\hat{\sigma}^2_{\psi_o}(\mathbf{z}_{m}, t_m)}\right]
\label{eqn:NLL} 
\end{equation}
,where $N_o$ denotes the total number of component instances of component class $o$ across all the $N$ samples, and $m$ denotes the index of the component instance in class $o$.

\textbf{Model Inference:} To estimate the CATE for a unit $i$, a modular architecture consisting of $m_i$ component models is instantiated with the same input and output structure as $G_i$. During inference for treatment $T=t$, the predictions of potential outcomes of the parent's components $\{\hat{y}_{lt}\}_{l \in Pa(c_j)}$, $l \in M_o$ are sampled from a normal distribution $\hat{y}_{lt} \sim N(\hat{\mu}_{ot}, \hat{\sigma}^{2}_{ot})$. We slightly abuse the notation to denote the inferred variance and variance model using $\sigma^2$. The mean and variance of the $j^{th}$ component outcome are obtained using these samples: $\hat{\mu}_{jt} = \hat{f}_{\theta^{*}_o}(\mathbf{x}_j, \{\hat{y}_{lt}\}_{l \in Pa(c_j)}, t) = \hat{f}_{\theta^{*}_o}(\hat{z}_{jt}, t)$,
$\hat{\sigma}^2_{jt} = \hat{\sigma}^2_{\psi^{*}_o}(\mathbf{x}_j, \{\hat{y}_{lt}\}_{l \in Pa(c_j)}, t)= \hat{\sigma}^2_{\psi^{*}_o}(\hat{z}_{jt}, t)$.
The estimate $\mathbb{E}[Y_i(1) - Y_i(0)|Q_i=q]$ of CATE is obtained through Monte Carlo integration by averaging over $S$ samples: $\hat{\tau}(q) \approx \frac{1}{S} \sum_{s=1}^S [\hat{f}_{\theta^{*}_o}(\hat{z}_{m_it}^{(s)}, 1) - \hat{f}_{\theta^{*}_o}(\hat{z}_{m_it}^{(s)}, 0)]$, where each sample path $\hat{z}_{m_it}^{(s)}$ is generated by hierarchically sampling from the component distributions: $\hat{y}_{jt}^{(s)} \sim \mathcal{N}(\hat{\mu}_{jt}, \hat{\sigma}_{jt}^2)$ for $j=1$ to $m_i-1$.

\subsection{Relaxing assumptions about component-level data access}
\label{sec:prior_knowledge_method}

The model description above assumes observed component-level covariates and outcomes. This assumption is often reasonable, given the wide availability of fine-grained data for many structured domains. However, other cases might exist when only the unit-level covariates $\mathbf{X}$ and outcomes are observed, and the component-level covariates $\mathbf{X}_j$ and outcomes $Y_j$ are unobserved. Below, we discuss hierarchical composition models for these cases. 

\textbf{Case 1: \textit{Unobserved} $\mathbf{X}_j$, \textit{observed} $Y_j$:}
We jointly learn the lower-dimensional component-level representations $\phi_o: \mathbb{R}^d \rightarrow \mathbb{R}^{d'_o}$, as well as the parameters of outcome functions $(\theta_o, {\psi_o})$ by assuming $z_j = (\phi_o(x_j),\{{y}_{lt}\}_{l \in Pa(c_j)}$) in Equation \ref{eqn:NLL}. 

\textbf{Case 2: \textit{Observed} $\mathbf{X}_j$, \textit{unobserved} $Y_j$: }The model architecture remains the same as before, but we do not have individual component-level loss functions and only know the loss function for unit-level outcomes. Due to this, the parameters of the components are jointly learned to optimize the loss of estimating unit-level outcomes. If $\circ$ denotes the functional composition, and functions are composed in the same hierarchical order as $G_i$, then the joint loss function is given by:
$[\theta_1, \theta_2 \dots \theta_k]:= \arg\min_{\mathbf{\Theta}} \frac{1} {N}\sum_{i=1}^{N} (\hat{f}_{\theta_{m_i}} \circ \hat{f}_{\theta_{m_i-1}} \circ  \dots \circ  \hat{f}_{\theta_{1}}(\mathbf{x}_{i1}, t_i)  - y_{i})^2$

\textbf{Case 3: \textit{Unobserved} $\mathbf{X}_j$ and \textit{unobserved} $Y_j$:} In this case, we only assume the knowledge of $G_i$. 
$[\theta_1, \theta_2 \dots \theta_k]:= \arg\min_{\mathbf{\Theta}} \frac{1} {N}\sum_{i=1}^{N} (\hat{f}_{\theta_{m_i}} \circ \hat{f}_{\theta_{m_i-1}} \circ  \dots \circ  \hat{f}_{\theta_{1}}(\phi_1(\mathbf{x}_i), t_i)  - y_{i})^2$

\textbf{Identifiability under non-observability:} Under the assumption of unobserved component-level data (e.g., case 3), the compositional model has unit-level information and structural knowledge, the same information as unitary models. The only difference between the two approaches is that structure is incorporated in the compositional model, while in the unitary model, it is passed in the form of high-level structural features. Thus, identification holds under the non-observability of component data access as long as the identifiability conditions (ignorability and overlap) hold for the unit-level covariates. As the unit-level features are aggregated, the ignorability might be affected due to the approximate representation of the units. 

\section{Experimental Infrastructure}
\label{sec:experiments}

We describe the experimental setup to evaluate the models across various \textit{in-distribution} and \textit{out-of-distribution} settings. We provide a detailed description of the causal effect estimation task, unit-level and component-level covariates, treatment, and outcomes for each domain in Section \ref{sec:experimental_infra_app} \footnote{Data and code for creating benchmarks and reproducing experiments is available at  \href{https://github.com/KDL-umass/compositional_models_cate}{https://github.com/KDL-umass/compositional\_models\_cate.}}.

\subsection{Datasets}
Research on modeling causal effects for compositional data has been hampered by the lack of real-world benchmarks. To facilitate effective empirical evaluation of the utility of compositional modeling of causal effect estimation, we introduce two benchmarks based on real-world computational systems and one benchmark based on a realistic simulation. 

\textbf{Query execution in relational databases:} We collected real-world query execution plans data by running 10,000 publicly available SQL queries against the Stack Overflow database under different configurations (memory size, indexing, page cost), treating configuration parameters as interventions and execution time as the potential outcome, assuming additive composition.

\textbf{Manufacturing plant data:} We use a discrete-event simulation framework, \href{https://simpy.readthedocs.io/en/latest/}{Simpy}, to generate realistic manufacturing plant data. The simulation includes four manufacturing processes (material processing, joining, electronics processing, and assembly) combined across 50 hierarchical manufacturing line layouts, assuming hierarchical composition. Each layout consists of various product demand (5-1,000) with different raw material inventories as covariates. Treatment compares five versus fifteen skilled workers, measuring total parts produced as outcomes.

\textbf{Matrix operations processing:} 
The dataset consists of 25,000 samples for 25 matrix expression structures (units) evaluated on two different computer hardware (treatment), with matrix dimensions ranging from 2 to 1,000. Component operations encompass $12$ component operations (e.g., multiplication, inverse, singular value decomposition, etc.). We ensure each operation is executed individually, ensuring parallel composition with additive aggregation function. Matrix size is used as a biasing covariate to create a distribution mismatch between treatment groups.

\textbf{Synthetic data:} In addition to these real-world benchmarks, we also generate simulated data to systematically understand the role of component-level data access, composition structure, and heterogeneity in components’ response surfaces. Composition structures include sequential and additive parallel composition. Structured units are generated by sampling binary trees (max depth=10) with $k=10$ heterogeneous modules, each having $d_j=1$ feature ($d=10$ features total). 

\subsection{Models and Evaluation Criteria}
\textbf{Compositional Models:} We implement four versions of the compositional models (hierarchical and parallel, depending on the domain) with varying amount of component-level data access as discussed in Sections \ref{section:hierarchical_composition_dist} and \ref{sec:prior_knowledge_method}: (1) \textit{Compositional (observed $\mathbf{X}_j$, observed $Y_j$)}; (2) \textit{Compositional (unobserved $\mathbf{X}_j$, observed $Y_j$)}; (3) \textit{Compositional (observed $\mathbf{X}_j$, unobserved $Y_j$)}; and (4) \textit{Compositional (unobserved $\mathbf{X}_j$, unobserved $Y_j$)}. The component-level models are implemented as neural networks, independently trained in case of access to fine-grained outcomes and jointly end-to-end trained otherwise. \textbf{Note:} Unless stated explicitly, the legend ``compositional" in experimental results implies a model with observed component-level covariates $\mathbf{X}_j$ and outcomes $Y_j$.

\textbf{Baselines:} We compare the performance of the compositional models with three types of existing approaches for CATE estimation: (1) \textit{TNet}, a neural network-based CATE estimator \citep{curth2021nonparametric}; (2) \textit{X-learner}, a meta learner that uses plug-in estimators to compute CATE, with random forest as the base model class \citep{kunzel2019metalearners}; (3) Non-parametric \textit{Double ML} \citep{chernozhukov2018double}; and (4) Vanilla \textit{neural network} and \textit{random forest}-based outcome regression models. 


\textbf{Creation of observational data sets:} Real-world computational systems provide experimental datasets with observed outcomes for both treatments, from which we create observational datasets by introducing structure and covariate-based confounding bias \citep{gentzel2021and}. Bias strength ranges from 0 (experimental) to 10 (observational), with treatment probabilities varying between 0.01-0.99, creating treatment group distribution shifts. Unconfoundedness holds as biasing information is observed in both approaches. Higher ``bias strength" indicates higher treatment probability for specific biasing covariate values and structure. 

\textbf{Evaluating compositional generalization:} We evaluate compositional generalization by training on structures with $2$ to $K$ module combinations and testing on combinations containing all $K$ module combinations.  For example, when training on 2-module combinations, we use all possible pairs (e.g., $(C_1, C_2)$, $(C_1, C_3)$, $(C_2, C_3)$), including varied orders and repeated modules (e.g., $(C_1, C_2, C_1)$), and test on larger combinations like $(C_1, C_2, C_3)$. For real-world data, we train on structures with smaller tree depths and test on larger ones.

\begin{figure}[h]
    \centering
    \includegraphics[width=\linewidth]{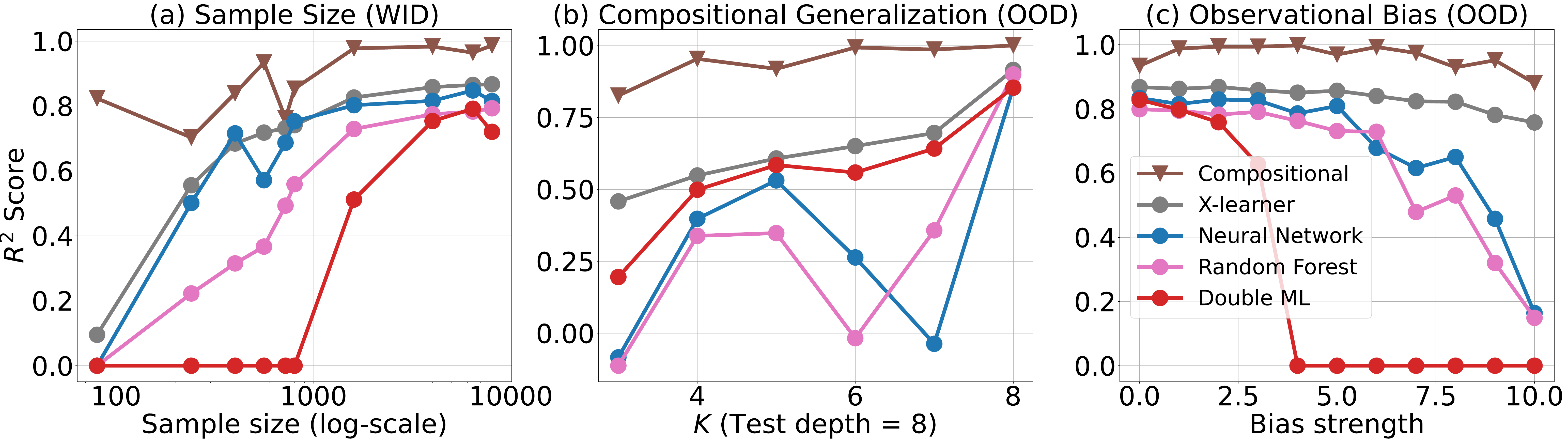}
    \caption{\textbf{Results on manufacturing domain} ($10,000$ samples). We report $R^2$ between CATE estimates and ground-truth effects (higher is better). (a) \textit{Sample-size efficiency (WID):} Compositional models are more accurate and sample-efficient in CATE estimation for within-distribution settings. (b) \textit{Compositional generalization (OOD):} Models are trained on units with tree depth $\leq K$ and evaluated on a test-set with depth $8$. Compositional models generalize to the unseen combinations due to compositional structure whereas non-compositional baselines perform comparably only after training on units similar to test data. (c) \textit{Effect of increasing observational bias (OOD):} Models are trained and tested on data with increasing observational bias strength between assigned treatment and tree depth. Compositional models and X-learner are less affected by increased observational bias.} 
    \label{fig:manufacturing_results}
\end{figure}

\textbf{Performance and Evaluation Metrics:} Performance is evaluated in two settings: \textit{WID} (same structure/covariate distribution in train/test) and \textit{OOD}, which includes (1) \textit{compositional generalization} (testing on unseen component combinations) and (2) \textit{observational bias} (structure/covariate-dependent treatment assignment). We measure PEHE loss \citep{hill2011bayesian}: $\epsilon_{PEHE}(\hat{f}) = \mathbb{E}[ (\hat{\tau}_{\hat{f}}(q) - \tau(q))^2]$, and $R^2$ score for datasets with large outcome values or high performance variability.

\section{Findings}
\label{sec:findings}

In this section, we provide the key findings from our experiments and discuss the mechanisms responsible for compositional models' performance compared to the baselines. 

\textbf{Compositional models can provide substantially more accurate CATE estimation for structured units:} Figure \ref{fig:manufacturing_results}(a) shows results from the manufacturing domain in which compositional and unitary models were learned from experimental data and evaluated in terms of their in-distribution performance. The compositional model has substantially lower error than the baselines, particularly for small sample sizes.  More detailed analysis showed that the performance advantage of compositional models in this setting is primarily due to two factors: (1) units in this domain have heterogeneous hierarchical structure and instance-specific models more accurately represent this structure; and (2) compositional models were learned with higher sample efficiency due to the existence of multiple samples of each component per unit (see Figure \ref{fig:comp_dist}(a) in the supplementary materials).

\begin{figure}[h]
    \centering
    \includegraphics[width=\linewidth]{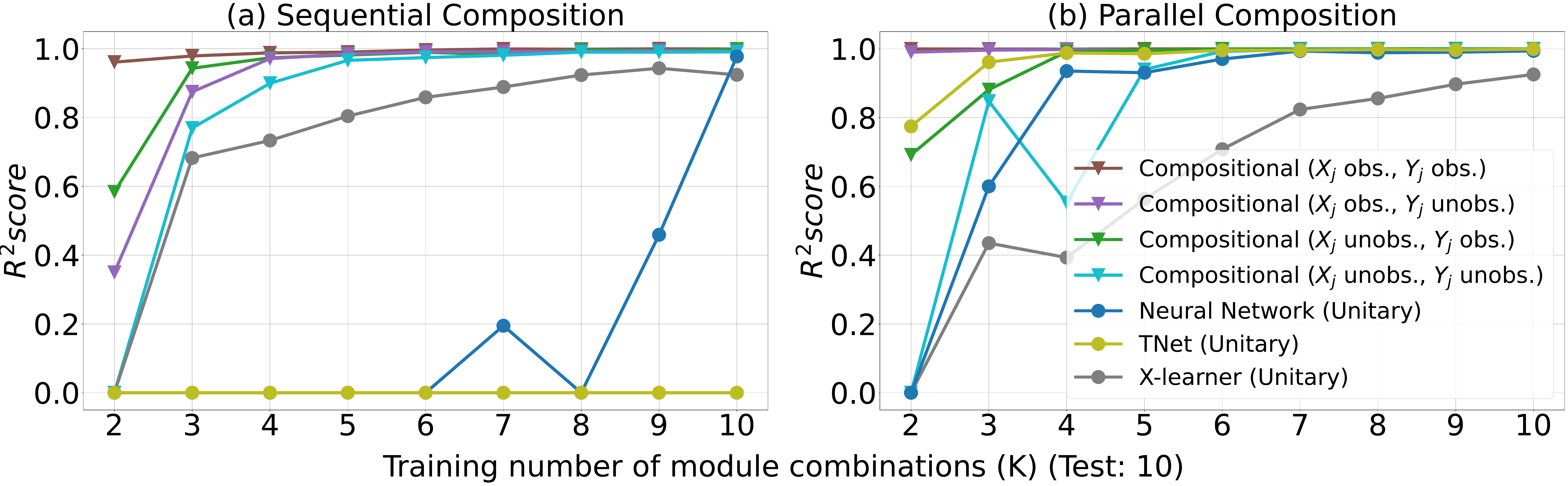}
    \caption{\textbf{Role of component-level data access and composition structure in the performance of compositional models:} $R^2$ score for models evaluated on compositional generalization task with varying degrees of component-level data access. (Higher is better; PEHE errors are reported in Figure \ref{fig:prior_knowledge_pehe_app} of the supplementary material). We observe that end-to-end trained models incorporating just modular structure compositionally generalize as trained on more module combinations. Unitary models show compositional generalization for additive parallel composition but perform comparably only for in-distribution combinations ($K$=$10$) for sequential composition, except X-learner. Note that the number of training samples increases as training depth increases.}
    \label{fig:prior_knowledge}
\end{figure}

\textbf{Incorporating modular structure enables compositional generalization:} Figures \ref{fig:manufacturing_results}(b) and \ref{fig:prior_knowledge} report the compositional generalization performance of the models in the manufacturing and synthetic domains, respectively. Compositional models were able to successfully generalize to high-complexity units (large depth or number of module combinations, respectively) even though they had only been trained on low-complexity units. In contrast, unitary models perform worse (often \textit{far} worse) until they are trained on units that equal or nearly equal the compositional complexity of the test units.


\begin{figure}
    \centering
    \includegraphics[width=\linewidth]{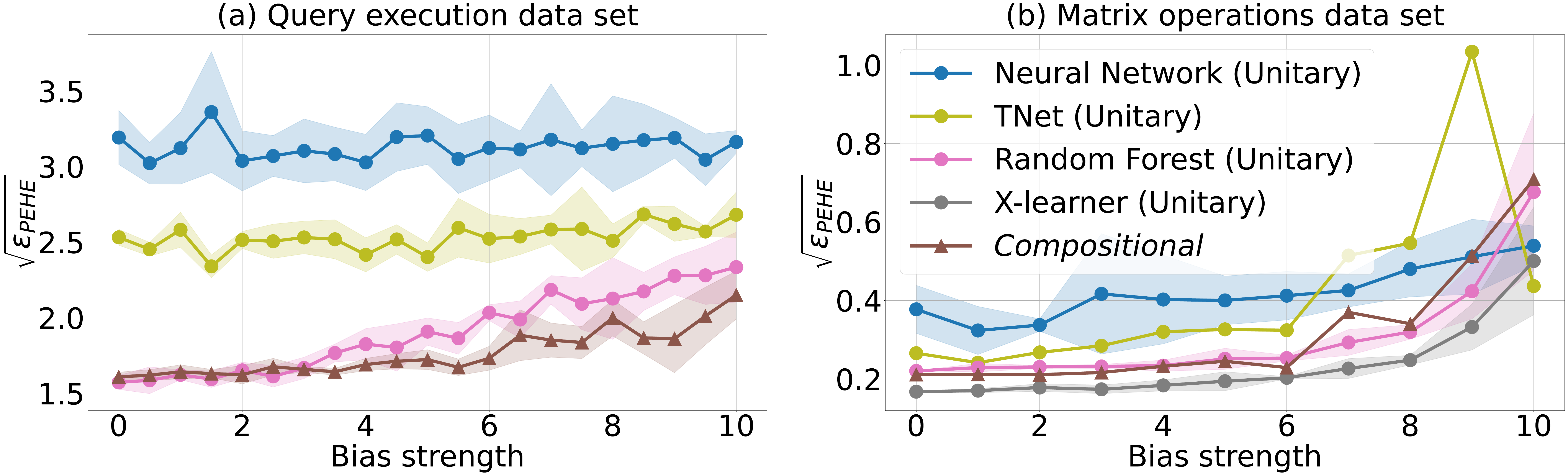}
    \caption{\textbf{Results for real-world data sets:} (a) \textit{Query execution data set:} Compositional model estimates the effect more accurately as observational bias increases. (b) \textit{Matrix operations data set:} All baselines perform similarly for this data set due to a single shared covariate, homogenous component outcome functions, and dominant contribution of matrix multiplication.}
    \label{fig:real_world_data_results}
\end{figure}

\textbf{Compositional models are less affected by observational bias in treatment assignment:}
Figure \ref{fig:manufacturing_results}(c) reports the effects of differing degrees of observational bias based on the instance-specific structure in the manufacturing domain. Compositional models (and the unitary X-learner) are least affected by this form of observational bias. Other unitary models (Neural network and Random forest) are strongly affected, and another unitary model (double ML) is the most strongly affected due to its use of propensity score weighting. Figure \ref{fig:real_world_data_results}(a) reports results for the query execution domain, where bias was introduced based on the covariate distribution. The error of all models increases because covariate-based bias affects both unit-level and component-level balance, but the error of the compositional model is lower and increases more slowly than the unitary baselines. 

\textbf{When trained end-to-end, compositional models perform remarkably well even without component-level data access:} Figure \ref{fig:prior_knowledge} reports the performance of models trained on synthetic data with varying degrees of data access. Figure \ref{fig:prior_knowledge}(a) and (b) report results on a synthetic domain in which units have sequential and parallel structures, respectively. Compositional models outperform unitary models regardless of their access to component-level covariates and outcomes (more for sequential case than parallel case as discussed below), demonstrating that compositional models can effectively estimate CATE for novel units in scenarios where component-level data is unavailable, and training uses only the compositional structure and unit-level data.

\textbf{Unitary approaches compositionally generalize more easily for additive composition structures than sequential structures:} The results in Figure \ref{fig:prior_knowledge} also show that unitary models can generalize for some composition types better than others. Figure \ref{fig:prior_knowledge}(a) and (b) show results from units simulated with strictly sequential and (additive) parallel composition, respectively, with exactly the same component functions and covariate distributions. While most unitary models perform very poorly in the case of sequential composition, nearly all perform fairly well in the case of additive parallel composition, particularly as module combinations become more similar to the test data. The role of structure type in compositional generalization is often overlooked in relevant work in machine learning, where most work assumes either additive or sequential compositions.

\textbf{Some factors can eliminate the advantages of compositional models for causal estimation:} Figure \ref{fig:real_world_data_results}(b) reports results for one realistic domain we studied --- matrix operations --- in which compositional models provide no substantial advantage over unitary baselines in both experimental (bias-strength = 0) and observational (bias-strength > 0) settings. This contrasts sharply with the superior performance of compositional models in the manufacturing (Figures \ref{fig:manufacturing_results}, \ref{fig:prior_knowledge}) and query execution (Figure \ref{fig:real_world_data_results}(a)) domains. Further investigation identified three factors that explain this result: (1) the \textit{dominance} of one component---matrix multiplication---in determining overall run-time (Figure \ref{fig:dominant_matmul}); (2) \textit{homogeneous} functional forms of the component outcome functions, such that additive composition leads to similar unit-level outcome functions (Figure \ref{fig:matrix_mult_plots}); and (3) a \textit{single shared covariate}---matrix size---that affects both unit-level and component-level outcomes, creating similar distribution imbalance issues at both the unit and component levels. In contrast, the query execution and manufacturing domains were more heterogeneous---different covariates affect component-level outcomes, no single component dominated in producing overall effects, component outcome functions belonged to different function classes, and composition structures were more complex. Thus, compositional models had higher relative performance in those domains.

\section{Conclusion}
The compositional models for causal effect estimation show promise in complex, modular, and heterogeneous systems. Compositional modeling provides a scalable and practical perspective for instance-specific causal reasoning in modern technological systems. 
This work focuses on compositional models for shared treatment and individual effect estimation. Compositional causal reasoning about component-specific treatments, such as selecting the optimal configuration parameters for each component, and reasoning about the wide array of realistic interventions, such as adding, replacing, or removing components from the system, are useful directions for future work. 
\acks{The authors thank Pracheta Amaranath, Andy Zane, Kate Avery, Sankaran Vaidyanathan, Justin Clarke, the anonymous reviewers and the area chair for helpful comments and suggestions. Thanks to Anirban Ghosh for helping design the manufacturing assembly simulator. This research was supported by DARPA and the United States Air Force under the  SAIL-ON (Contract No. w911NF-20-2-0005) program. Purva also received the Dissertation Writing Fellowship from the Manning College of Information and Computer Sciences in the Spring of 2024. Any opinions, findings, conclusions or recommendations expressed in this document are those of the authors and do not necessarily reflect the views of DARPA, ARO, the United States Air Force, or the U.S. Government. The U.S. Government is authorized to reproduce and distribute reprints for Government purposes notwithstanding any copyright notation herein.} 

\bibliography{references}

\newpage

\appendix

\section{Other examples of structured systems with compositional data}
\label{sec:real_world_examples}
The causal questions of interest in the compositional domain are: How do the unit-level interventions impact the component-level outcomes to produce the overall unit’s outcome? Many real-world phenomena require answering such causal questions about the effect of shared interventions on different components. We provide several real-world use cases where the compositional approach can be useful to reason about the effects of the interventions and make informed and personalized decisions. 
\begin{itemize}
    \item \textit{Compiler optimization:} How do different hardware architectures \textit{(intervention)} affect the compile time \textit{(potential outcome)} of different source codes \textit{(unit)} ? In this case, source code consists of multiple program modules; hardware architecture is the unit-level intervention that can affect the compiling of different source codes differently, and compile time is the outcome of interest. 
    \item \textit{Energy efficiency optimization:} How does a state-wide mandate \textit{(intervention)} of shifting to more efficient electric appliances affect the monthly bill \textit{(potential outcome)} of each building \textit{(unit)} in the state? Each building can be assumed to consist of various compositions of electric appliances. The intervention might affect the bill of each kind of appliance differently, affecting the overall utility bill.
    \item \textit{Supply chain optimization:} How is the processing time \textit{(potential outcome) }of an order \textit{(unit)} affected when a supply chain company shifts to a different supplier for various parts \textit{(intervention)}? In this case, each order execution plan is the unit of analysis that consists of routing the information from different parties, suppliers, manufacturers, and distributors specific to each order; intervention can impact the processing time of different parties depending on the affected parts and order details.
    \item \textit{Causal reasoning in multi-agent systems:} How do the model hyperparameters (architecture, agent implementation types) affect the accuracy (potential outcomes) of multi-agent systems (LLM agents) for different task instances (unit)? As multi-agent LLM-based systems are becoming an integral part of daily workflows, developing instance-specific causal reasoning models for such systems using compositional approaches is helpful. 
    
\end{itemize}

\section{Related Work}
\label{sec:related work}
In this section, we discuss the connections of the compositional approach with the existing work in causal inference and associational machine learning in greater detail. 

\textbf{Causal effect estimation in structured domains:} In causal inference, a relatively sparse body of work has focused on treatment effect estimation on structured data in modular domains \citep{gelman2006data, salimi2020causal, kaddour2021causal}. For example, existing work in multi-level modeling and hierarchical causal models \citep{gelman2006data, witty2018causal, weinstein2024hierarchical} leverages hierarchical data structure to improve effect estimation under unobserved confounders. There is also growing interest in heterogeneous effect estimation for complex data, such as images \citep{jerzak2023image}, structured treatments (e.g., graphs, images, text, drugs) \citep{harada2021graphite, kaddour2021causal}, and relational data \citep{salimi2020causal, khatami2024graph}. The compositional approach complements this line of research by focusing on the units composed of multiple heterogeneous components. On the other hand, hierarchical causal models focus on units with hierarchical structures but homogeneous sub-units in a unit. Relational causal models primarily employ relational semantics to describe instance-specific structures and interactions among entities. The compositional approach uses simpler compositional semantics and focuses on the generic component-wise behavior of a system and interactions among them, where components can be objects, processes, and distinct heterogeneous modules in a system.  Our focus also lies in the structured and compositional representation of the units rather than only treatments, which helps better estimate causal effects in the case of high-dimensional observational data. Another piece of related work is the fine-grained analysis of the potential outcomes to study the validity of synthetic control methods with panel data \citep{shi2022assumptions}. This work focuses on reasoning about potential outcomes of homogeneous sub-units (individuals) to establish identifiability of unit-level (state) potential outcomes. The compositional approach employs similar fine-grained reasoning, but explicitly uses the fine-grained data for compositional causal modeling. 

\textbf{Modularity and compositionality in SCMs:} The vast body of work under the structural causal model (SCM) framework \citep{pearl2009causality} typically summarizes a system’s behavior with a fixed set of variables and assumes fixed causal structure among those variables (unless modified under explicit intervention via the do-operator). The causal model, a directed acyclic graph, represents the causal interactions among all variables. In the compositional approach, we assume an instance-specific structure among the components for a given unit. Thus, the compositional data-generating process could not be represented by a single SCM. Instead, a unique SCM corresponding to each composition structure would be required.  This is beyond the scope of nearly all current work in SCMs, with a very few exceptions (e.g., \citep{laskey2008mebn}).
 
In the specific context of structural causal models, the term \textit{modularity} is sometimes used to refer to a model property in which the structural function of a given variable can be intervened upon without influencing the structural function of any other variable. This property is also known as \textit{autonomy} \citep{haavelmo1944probability}, \textit{structural invariance} \citep{aldrich1989autonomy}, and \textit{independence of causal mechanism} \citep{peters2017elements}. Note that modularity, in this sense, is a property of the model--- It is true by definition (variables in an SCM are assumed to be modular), and it is absolute. In contrast, the modular structure that we reference in this paper is a property of both the system being modeled and (perhaps) the structure of a given model of that system. To enable compositional generalization, the compositional approach assumes that the data-generating distribution of outcomes of a component given component-specific inputs remains stable and invariant across the units \textit{irrespective} of where the component appears in the structure of a unit. This assumption differs from the modularity assumption made in SCM, which assumes the stability for the conditional distribution of \textit{each} random variable given its parents in the direct graph with respect to the interventions on the other conditional distributions.  

\textbf{Compositional models in associational machine learning:} Our work is inspired by research on compositional models in machine learning that exploit the structure of underlying domains and explicitly represent it in the model structure \citep{heckerman1995bayesian, koller1997object, friedman1999learning, getoor2007introduction, taskar2005learning, laskey2008mebn}. For example, research in object-oriented and relational models has produced directed graphical models that explicitly reproduce known modular structure in the systems being analyzed \citep{koller1997object, friedman1999learning, laskey2008mebn}. In a similar way, modular architectures in deep neural networks have been designed to replicate the assumed modular structure of the systems that they attempt to model \citep{jacobs1991adaptive}. Some work in probabilistic programming has a similar flavor, in that the structure of the probabilistic program reflects known modular structure in the real-world system being analyzed \citep{lake2015human}. The \textit{instance-specific} modular architectures are widely used in machine learning to model data in natural language, program synthesis, reinforcement learning, and combined vision and language problems, providing sample efficiency, systematic generalization, and computation benefits \citep{andreas2016neural, shazeer2016outrageously, pfeiffer2023modular}. The closest work to the compositional models for causal effect estimation is using a mixture of expert (MoE) architecture \citep{jacobs1991adaptive, shazeer2016outrageously} and modular neural networks \citep{socher2011parsing, andreas2016neural, marcus2019plan} in vision and language domains. However, most of the work in machine learning focuses on understanding the systematic generalization and sample efficiency benefits of compositional models for prediction tasks. At the same time, their role in reasoning about intervention effects is unexplored \citep{lake2018generalization, hupkes2020compositionality, mittal2022modular, jarvisspecialization, wiedemer2024compositional, schugdiscovering, lippl2024impact}. We inspire the formalization of the compositional data-generating process using a systems perspective (e.g., query execution system) with explicit causal mechanisms and interventions. In contrast, vision/language domains often lack well-defined components and causal interactions due to the perceptual nature of the data. Additionally, compositionality is usually a property of the causal data-generating process, and viewing compositionality from the lens of causality, interventions, and components in the modular systems helps understand the compositional generalization characteristics of the models. This work takes the first step in that direction. 

\section{Graphical representation of compositional data-generating process}
\label{sec:plate_based_notation}
In section \ref{sec:compositional_framework}, we use the potential outcomes (PO) framework \citep{rubin1974estimating, rubin2005causal} to describe the compositional data generating process for causal inference because the PO framework simplifies the representation of the key elements of the compositional approach: (1) \textit{small number of causal dependencies:} our focus is on estimating the causal effect of treatment (T) on component outcomes ($Y_j$) given component covariates ($\mathbf{X}_j$) without needing to explicitly model causal relationships among the covariates (and assuming that causal mechanisms of all random variables are modular), as required by structural causal models; (2) \textit{component interactions:} we represent component interactions and composition structure through the interaction graph (Figure 1(a),(c)). While $G_i$ constrains the possible causal dependencies among random variables that could be represented as instance-specific causal graphs, $G_i$ also highlights the most important aspect of compositional models: the manner in which components interact to bring about system behavior. These elements are formalized through Equation \ref{eqn:Markov_dependence}. 

\begin{figure}[h]
    \centering
\includegraphics[width=\linewidth]{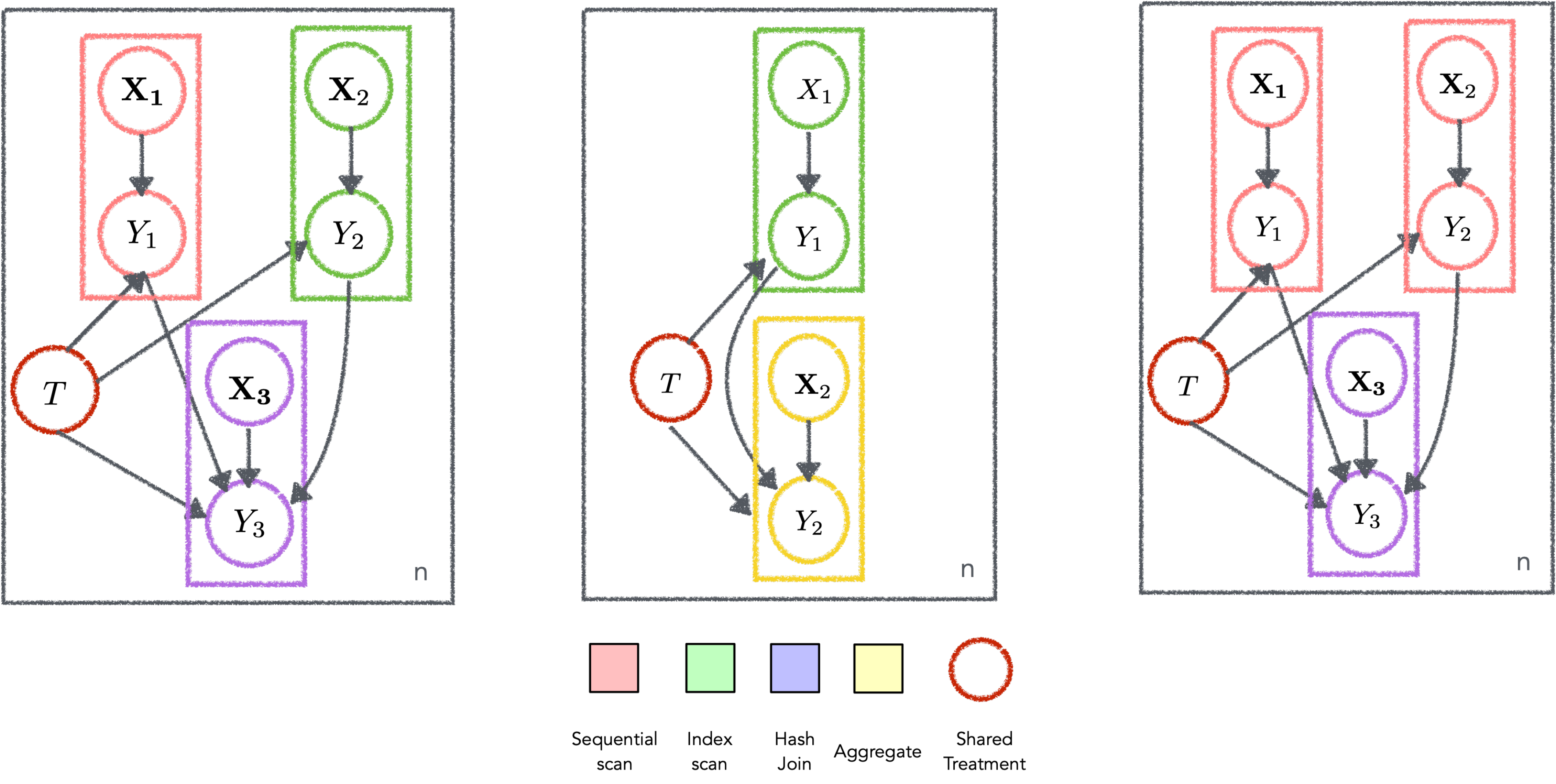}
    \caption{\textbf{Graphical representation of compositional causal models:} Each plate model represents the data-generating process of unit-level and component-level variables for a given instance-specific composition structure ($G_i$), shown for three different structures here. $X_j$ denotes the component-specific covariates, $Y_j$ denotes the component-specific outcomes, and $T$ denotes the unit-level shared treatment. Each distinct color represents the fixed data generating process for a specific component, that might appear in multiple units. }
    \label{fig:nested_plate_notation}
\end{figure}

Compositional data generating process described in Section \ref{sec:compositional_framework} can also be represented using a plate-based notation, commonly used in the probabilistic graphical models to represent various model classes such as hierarchical causal models \citep{gelman2006data, witty2018causal, weinstein2024hierarchical} and relational causal models \citep{maier2013sound, lee2016learning, salimi2020causal, ahsan2023learning}. In Figure \ref{fig:nested_plate_notation}, we represent plate models for three different composition structures in which component-level variables lie inside an inner plate and the unit-level variables lie in the outer plate, with edges denoting the causal dependencies among component-specific covariates, potential outcomes, and unit-level treatment.   

\section{Identifiability Results }

\subsection{Definition, assumptions and auxiliary lemmas}
 We first define the necessary distributions and provide some simple results. Assume that each unit $i$ has pre-treatment covariates $\mathbf{X}_i = \mathbf{x} \in \mathcal{X} \subset \mathbb{R}^d$, a binary treatment $T_i \in \{0,1\}$, and two potential outcomes $\{Y_i(0), Y_i(1)\}\in \mathcal{Y} \subset \mathbb{R}$ \citep{rubin1974estimating, rubin2005causal}. In the observational data, we only observe one of the potential outcomes for each unit, $Y_i  = Y_i(T_i)$ known as the \textit{observed} or \textit{factual} outcome and missing outcomes ${Y_i}_{CF} = Y_i(1-T_i)$ are known as \textit{observed} or \textit{counterfactual} outcome. 

\begin{definition}
    The conditional average treatment effect (CATE) is defined as 
    $$\tau(x): \mathbb{E}[Y_i(1) - Y_i(0) | \mathbf{X}_i = \mathbf{x}]$$
\end{definition}
We first show that under the assumptions of ignorability and consistency, the CATE function $\tau(x)$ is identifiable by $\tau(\mathbf{x}) = \mathbb{E}[{Y_i|\mathbf{X}_i = \mathbf{x}, T=1}] - \mathbb{E}[{Y_i|\mathbf{X}_i=\mathbf{x}, T=0}]$ \citep{pearl2009causality} \citep{rosenbaum1983central}. We assume a joint distribution function $p(\mathbf{X}_i, T_i, Y_i(1), Y_i(0))$, such that $(Y_i(1), Y_i(0) \perp T_i)|\mathbf{X}_i$ and $0 < P(T=1|\mathbf{X}_i=\mathbf{x}) < 1$, for all $x$. We also assume consistency; that is, we assume that we observe $y_i = Y_i(1)|T_i=1$ and $y_i = Y_i(0)|T_i=0$.
\begin{lemma}
    $$\mathbb{E}[Y_i(1) - Y_i(0) | \mathbf{X}_i = \mathbf{x}]$$
    \begin{equation}
    \label{app:cate_linearity}
        =\mathbb{E}[Y_i(1)|\mathbf{X}_i = \mathbf{x}] - \mathbb{E}[Y_i(0) | \mathbf{X}_i = \mathbf{x}]
    \end{equation}
    \begin{equation}
    \label{app:cate_ignorability}
        =\mathbb{E}[Y_i(1)|\mathbf{X}_i = \mathbf{x}, T_i=1] - \mathbb{E}[Y_(0) | \mathbf{X}_i = \mathbf{x}, T_i=0]    
    \end{equation}
    \begin{equation}
    \label{app:cate_consistency}
        =\mathbb{E}[Y_i|\mathbf{X}_i = \mathbf{x}, T_i=1] - \mathbb{E}[Y_i| \mathbf{X}_i = \mathbf{x}, T_i=0]    
    \end{equation}
\end{lemma}

Equality (\ref{app:cate_linearity}) is due to the linearity of the expectation. Equality (\ref{app:cate_ignorability}) follows from the ignorability assumption in which we assume that $Y(T)$ is independent of $T$ conditioned on $\mathbf{X}$. Equality (\ref{app:cate_consistency}) follows from the consistency assumption. The last equation consists of only observable quantities and can be estimated from the data if we assume overlap, $0 < P(T_i=1|\mathbf{X}_i = \mathbf{x}) < 1$, for all $\mathbf{x}$.

\begin{definition}
The conditional-average treatment effect (CATE) estimand for structured input $Q_i=q$ is defined as: $\tau(q) = \mathbb{E}[Y_i(1) - Y_i(0) | Q_i = q] = \mathbb{E}[Y_i(1) - Y_i(0) | Q_i = (G_i, \{\mathbf{x}_{ij}\}_{j = 1:m_i})]$
\end{definition}
For ease of notation to describe conditional distributions, we denote the combined inputs to a component $j$ as $\mathbf{Z}_j(t) = (\mathbf{X}_j, \{Y_l(t)\}_{l \in {Pa(c_j)}})$.

\begin{definition}
    The conditional-average treatment effect (CATE) estimand for component $j$ with $\mathbf{X}_j = \mathbf{x}_j \in \mathbb{R}^{d_j}$ is defined as: $\tau(\mathbf{z}_j) = \mathbb{E}[Y_i(1) - Y_i(0) | \mathbf{Z}_j = \mathbf{z}_j]$.
\end{definition}

\subsection{Identifiability for hierarchical composition}

We define the component-wise  distributions as $P(Y(t)|\mathbf{Z}_j(t) = \mathbf{z}_j)$. In hierarchical composition, the unit outcome equals the final component outcome: $\mathbb{E}[Y_i(t) |Q_i = q] = \mathbb{E}[Y_{im_i}(t)|Q_i=q]$. This conditional expectation can be expressed by marginalizing intermediate component outcomes using the Markov assumption (Equation \ref{eqn:Markov_dependence}).
\begin{equation}
    \mathbb{E}[Y_i(t) |Q_i = q] = \int_{Y_{im_i-1}(t)} \int_{Y_{im_i-2}(t)} \dots \int_{Y_{i1}(t)}\mathbb{E}[Y_{im_i}(t)|\mathbf{Z}_{im_i}(t)] \prod_{j=1}^{m_i-1} P(Y_{ij}(t)|\mathbf{Z}_{ij}(t))
\label{eqn:hierarchical_factorization}
\end{equation}

We use the following nested expectation expression as shorthand for marginalization over intermediate component outcomes: 
\begin{equation}
    \mathbb{E}[Y_i(t) |Q_i = q] = \mathbb{E}_{Y_{i1:im_i-1}(t)}[\mathbb{E}[Y_{im_i}(t)|\mathbf{Z}_{im_i}(t)]]
\label{eqn:hierarchical_factorization_shorthand}
\end{equation}

\begin{assumption}[Component-level ignorability]
The component level potential outcomes and assigned treatment are independent conditioned on the component level covariates, i.e.,  $Y_{j}(1), Y_{j}(0) \perp T | \mathbf{X}_{j}$. 
\label{assumption:c_ignorability}
\end{assumption}

\begin{assumption}[Component-level overlap]
The overlap holds for the component level covariates $X_j = x_j$, i.e., 
$\forall x_j \in \mathcal{X}_j, t \in \{0,1\}: \  0 < p(T=t|\mathbf{X}_{j}=\mathbf{X}_{j}) < 1$
\label{assumption:c_overlap}
\end{assumption}
\begin{assumption}[Component-level consistency]
The consistency holds for the component level covariates, i.e.,  $y_{j} = Y_{j}(0)|t=0$ and $y_{j} = Y_{j}(1)|t=1$.
\label{assumption:c_consistency}
\end{assumption}
As the treatment is assigned before any component’s potential outcomes are observed, we can assume that the component’s covariates $\mathbf{X}_j$ are sufficient to satisfy ignorability assumptions, i.e., $Y_{j}(t) \perp T |X_{j}$
Consider the conditional distribution $P(Y_{j}(t)|\mathbf{Z}_{j}(t))$. Assuming component-level ignorability for component $j$, we can write

$$P(Y_{ij}(t)|\mathbf{Z}_{ij}(t)) = P(Y_{ij}(t)|\mathbf{X}_j, \{Y_l(t)\}_{l \in {Pa(c_j)}})$$
$$ = P(Y_{ij}|\mathbf{X}_j, \{Y_l(t)\}_{l \in {Pa(c_j)}}, T = t)$$

Now, assuming consistency for components $c_l$, where $l \in Pa(c_j)$, we can write $Y_l(t)|T=t = Y_l$.
$$P(Y_{ij}(t)|\mathbf{Z}_{ij}(t)) = P(Y_{ij}|\mathbf{X}_j, \{Y_l\}_{l \in {Pa(c_j)}}, T = t) =  P(Y_{ij}|\mathbf{Z}_{ij}, T=t) $$

Similarly, $\mathbb{E}[Y_{im_i}(t)|\mathbf{Z}_{im_i}(t)]$ can be written as $\mathbb{E}[Y_{im_i}|\mathbf{Z}_{im_i}, T = t]$, assuming component-level ignorability and consistency. Substituting these quantities in Equation \ref{eqn:hierarchical_factorization}, we get the below result. 

$$\tau(q) = \int_{Y_{im_i-1}} \int_{Y_{im_i-2}} \dots \int_{Y_{i1}} \mathbb{E}[Y_{im_i}|\mathbf{Z}_{im_i} = \mathbf{z}_{im_i}, T = 1] \prod_{j=1}^{im_i-1} P(Y_{ij}|\mathbf{Z}_{ij} = \mathbf{z}_{iz}, T= 1) - $$

$$\int_{Y_{im_i-1}} \int_{Y_{im_i-2}} \dots \int_{Y_{i1}} \mathbb{E}[Y_{im_i}|\mathbf{Z}_{im_i} = \mathbf{z}_{im_i}, T = 0] \prod_{j=1}^{im_i-1} P(Y_{ij}|\mathbf{Z}_{ij} = \mathbf{z}_{iz}, T= 0)$$

Using shorthand notation, we get the desired result. 
 $$\tau(q) = \mathbb{E}_{Y_{1:m_i-1}}[\mathbb{E}[Y_{m_i}|\mathbf{Z}_{m_i} = \mathbf{z}_{m_i}, T = 1]] - \mathbb{E}_{Y_{1:m_i-1}}[\mathbb{E}[Y_{m_i}|\mathbf{Z}_{m_i} = \mathbf{z}_{m_i}, T = 0]]$$
Component-level overlap assumption ensures that the estimand is identified using observational data. 

\subsection{Identifiability for additive parallel composition}

In this section, we first describe the assumptions that the data-generating process follows for additive parallel composition. Then, we prove the identifiability results for additive parallel composition.

\begin{assumption}
\textbf{Additivity} assumes that the ground-truth component-level potential outcomes add up to generate the ground-truth unit-level potential outcome, i.e., $Y_i(1) = \sum_{j=1}^{m_i} Y_{ij}(1)$, $Y_i(0) = \sum_{j=1
}^{m_i} Y_{ij}(0)$.
\label{assumption:additivity_app}
\end{assumption}

\begin{assumption}
    \textbf{Conditional independence} assumption among potential outcomes implies that the ground-truth potential outcomes of a component $j$ are conditionally independent of outcomes of other components l ($l \neq j$) given the component's covariates $\mathbf{X}_j$: $Y_j(T) \perp Y_l(T) |  \mathbf{X}_j$
\label{assumption:conditional_independence_app}
\end{assumption}

Assuming conditional independence assumption and Markov assumption, the data-generating process for additive parallel composition can be written as:
$Y_{ij}(t) = \mu_{ot}(\mathbf{X}_{ij}) + \epsilon_{io}(t)$

Assuming additivity assumption \ref{assumption:additivity_app}, we get 
$$\tau(q) = \mathbb{E}[\sum_{j=1}^{m_i} Y_{ij}(1) - \sum_{j=1}^{m_i} Y_{ij}(0) | Q_i = (G_i, \{\mathbf{x}_{ij}\}_{j = 1:m_i})]$$
Due to the linearity of the expectation, we get the following: 
$$\tau(q) = \mathbb{E}[\sum_{j}^{m_i} Y_{ij}(1)| Q_i = (G_i, \{\mathbf{x}_{ij}\}_{j = 1:m_i})] - \mathbb{E}[\sum_{j=1}^{m_i} Y_{ij}(0) | Q_i = (G_i, \{\mathbf{x}_{ij}\}_{j = 1:m_i})]$$

Assuming conditional independence assumption among the component-level potential outcomes and Markov assumption, we get
$$\tau(q) = \sum_{j=1}^{m_i} \mathbb{E}[Y_{ij}(1)| \mathbf{x}_{ij}] - \mathbb{E}[Y_{ij}(0) | \mathbf{x}_{ij}] = \sum_{j=1}^{m_i} \mathbb{E}[Y_{ij}(1) - Y_{ij}(0) | \mathbf{x}_{ij}]= \sum_{j=1}^{m_i} \tau(\mathbf{x}_{ij}) $$

Now, we prove the identifiability result for additive parallel composition.

$$\tau(q) = \sum_{j=1}^{m_i} \mathbb{E}[Y_{ij}(1)| \mathbf{x}_{ij}] - \mathbb{E}[Y_{ij}(0) | \mathbf{x}_{ij}]$$
Assuming component-level ignorability, we get
$$\tau(q) = \sum_{j=1}^{m_i} \mathbb{E}[Y_{ij}(1)| \mathbf{x}_{ij}, T = 1] - \mathbb{E}[Y_{ij}(0) | \mathbf{x}_{ij}, T = 0]$$ Assuming component-level consistency, we get
$$\tau(q) = \sum_{j}^{m_i} \mathbb{E}[Y_{ij}| \mathbf{x}_{ij}, T = 1] - \mathbb{E}[Y_{ij}| \mathbf{x}_{ij}, T = 0]$$
Component-level overlap assumption ensures the estimand is identified using observational data.

\section{Learning compositional models from observational data}
\label{section:composition_models_app}

In this section, we discuss the algorithm for the additive parallel composition model discussed in Section 2.4. 

\begin{figure}[h]
    \centering
    \includegraphics[width=0.5\linewidth]{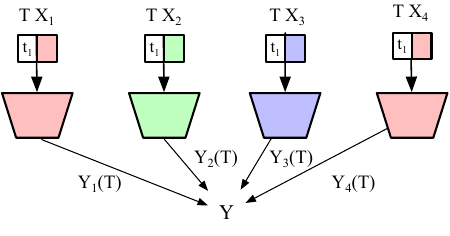}
    \caption{Model architecture for parallel composition model}
    \label{fig:additive_parallel}
\end{figure}
\label{section:additive_parallel_composition_model}
The key idea is that the component-level models for effect estimation are instantiated specific to each unit, and outcomes of one component are not shared with other components as we assume conditional independence among the potential outcomes given component-level features and shared treatment. In addition, linearity of expectation applies to the additive composition model, so we can directly compute the CATE estimates by just estimating the expected component-level potential outcomes. See Figure \ref{fig:additive_parallel} for model architecture for parallel composition.

\textbf{Model Training:} The component models for estimating component-level potential outcomes are denoted by $\hat{f}_{\theta_o}: \mathbb{R}^{d_o} \times \{0,1\} \rightarrow \mathbb{R}$. Each model corresponding to component class $o \in \{1, 2, \dots k\}$ is parameterized by separate and independent parameters $\theta_o$. For a given observational data set with $n$ samples, $\mathcal{D}_F = \{q_i, t_i, y_i\}_{i=1:n}$, we assume that we observe component-level features $\{\mathbf{x}_{ij}\}_{j=1:m_i}$, assigned treatment $t_i$ and fine-grained component-level potential outcomes $\{y_{ij}\}_{j=1:m_i}$ along with unit-level potential outcomes $y_i$. If component instance $c_j \in M_o$, training of each component model $o$ involves the independent learning of the parameters by minimizing empirical risk: $\theta^{*}_o:= \arg\min_{\theta_o} \frac{1} {N_o}\sum_{m=1}^{N_o} (\hat{f}_{\theta_o}(\mathbf{x}_{m}, t_m) - y_{m})^2$, where $N_o$ denotes the total number of component instances of component class $o$ across all the $N$ samples, and $m$ denotes the index of the component instance belonging to class $o$. To clarify, $t_m$ denotes the assigned treatment $T_i$ for the unit $i$ from which component instance $m$ data sample is obtained, and $y_m$ is the corresponding component-level outcome. The potential outcome of each component is computed using input features of that component, shared unit-level treatment, and \textit{observed} potential outcomes of the parent's component. 


\textbf{Model Inference:} To estimate CATE for a unit $i$, a modular architecture consisting of $m_i$ component models is instantiated with the same number of components as in the unit $i$. During inference for treatment $T=t$, due to conditional independence assumption, $\hat{y}_{ijt} = \hat{f}_{\theta^{*}_o}(\mathbf{x}_{ij}, t)$.
The estimate of CATE is obtained by taking the sum of the potential outcome estimates of all component instances $\hat{\tau}(q) = \sum_{j=1, j \in M_o}^{m_i}  \hat{f}_{\theta^{*}_o}(\mathbf{x}_{ij}, 1) - \hat{f}_{\theta^{*}_o}(\mathbf{x}_{ij}, 0)$. 

\subsection{Relaxing assumptions about component-level data access for additive parallel composition}
The model description above assumes observed component-level covariates and outcomes. This assumption is often reasonable, given the wide availability of fine-grained data for many structured domains. However, other cases exist when only the unit-level covariates $\mathbf{X}$ and outcomes are observed, and the component-level covariates $\mathbf{X}_j$ and outcomes $Y_j$ are unobserved. Below, we discuss hierarchical composition models for these cases.

\textbf{Case 1: \textit{Unobserved} $\mathbf{X}_j$, \textit{observed} $Y_j$:}
We jointly learn the lower-dimensional component-level representations $\phi_o: \mathbb{R}^d \times \mathbb{R}^{d'_o}$, as well as the parameters of outcome functions. If we assume  component instance $c_j \in M_o$, then $\hat{f}_{\theta_o}$:  $\theta_o:= \arg\min_{\theta} \frac{1} {N_o}\sum_{i=1}^{N_o} (\hat{f}_{\theta_o}(\phi_o(\mathbf{x}_i), t_i) - y_{ij})^2$. $\phi_o$ is jointly trained with the parameters $\theta_o$, so that the relevant variables for predicting $y_{ij}$ are selected. 

\textbf{Case 2: \textit{Observed} $\mathbf{X}_j$, \textit{unobserved} $Y_j$: }The model architecture remains the same as before, but we do not have individual component-level loss functions and only know the loss function for unit-level outcomes. Due to this, the parameters of the components are jointly learned to optimize the loss of estimating unit-level outcomes. Due to additive composition, the joint loss function is given by:
$[\theta_1, \theta_2 \dots \theta_k]:= \arg\min_{\mathbf{\Theta}} \frac{1} {N}\sum_{i=1}^{N} (\hat{f}_{\theta_{m_i}} + \hat{f}_{\theta_{m_i-1}} +  \dots +  \hat{f}_{\theta_{1}}(\mathbf{x}_{i1}, t)  - y_{i})^2$

\textbf{Case 3: \textit{Unobserved} $\mathbf{X}_j$ and \textit{unobserved} $Y_j$:} In this case, we only assume the knowledge of $G_i$. In this case, the model is equivalent to a mixture of experts (MoE) \citep{jacobs1991adaptive} architecture with addition as gating function. 
$[\theta_1, \theta_2 \dots \theta_k]:= \arg\min_{\mathbf{\Theta}} \frac{1} {N}\sum_{i=1}^{N} (\hat{f}_{\theta_{m_i}} + \hat{f}_{\theta_{m_i-1}} + \dots +  \hat{f}_{\theta_{1}}(\phi_1(\mathbf{x}_i), t)  - y_{i})^2$

\section{Experimental Infrastructure}
\label{sec:experimental_infra_app}

\begin{figure}[t]
    \centering
    \includegraphics[width=\linewidth]{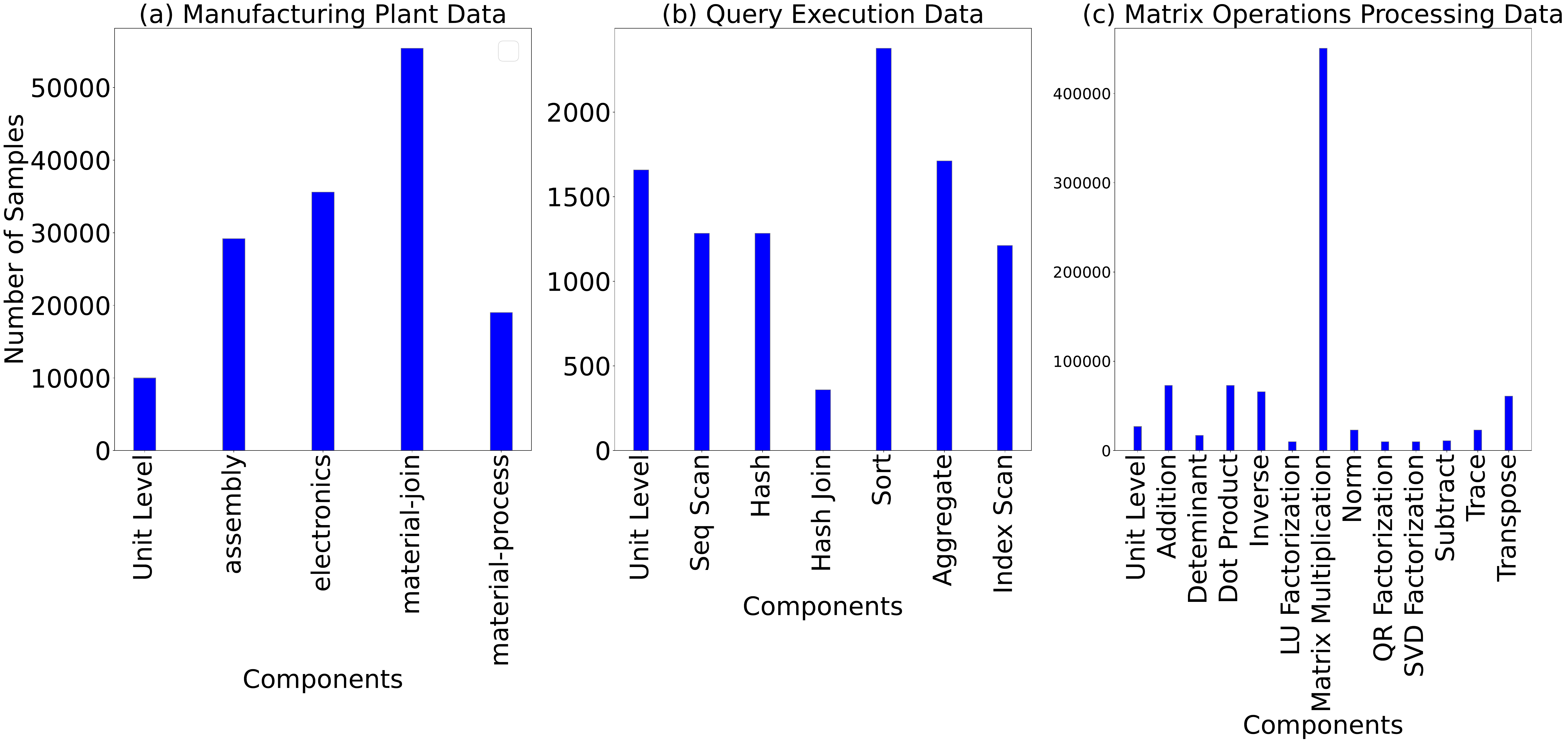}
    \caption{\textbf{Number of samples for units and component instances for different domains:} (a) Manufacturing data set showing component re-use across layouts ($10,000$ units). (b) Query execution data set  ($1,500$ units). (c) Matrix operations processing data set ($25,000$ units)}
    \label{fig:comp_dist}
\end{figure}

\subsection{Data sets}
In this section, we describe the details of the two benchmarks based on real-world computational systems --- query execution, matrix operations and one benchmark based on a realistic simulation --- manufacturing plant data.
\subsubsection{Query Execution System}

\begin{figure}
\vskip 0.2in
\centerline{\includegraphics[width=\columnwidth]{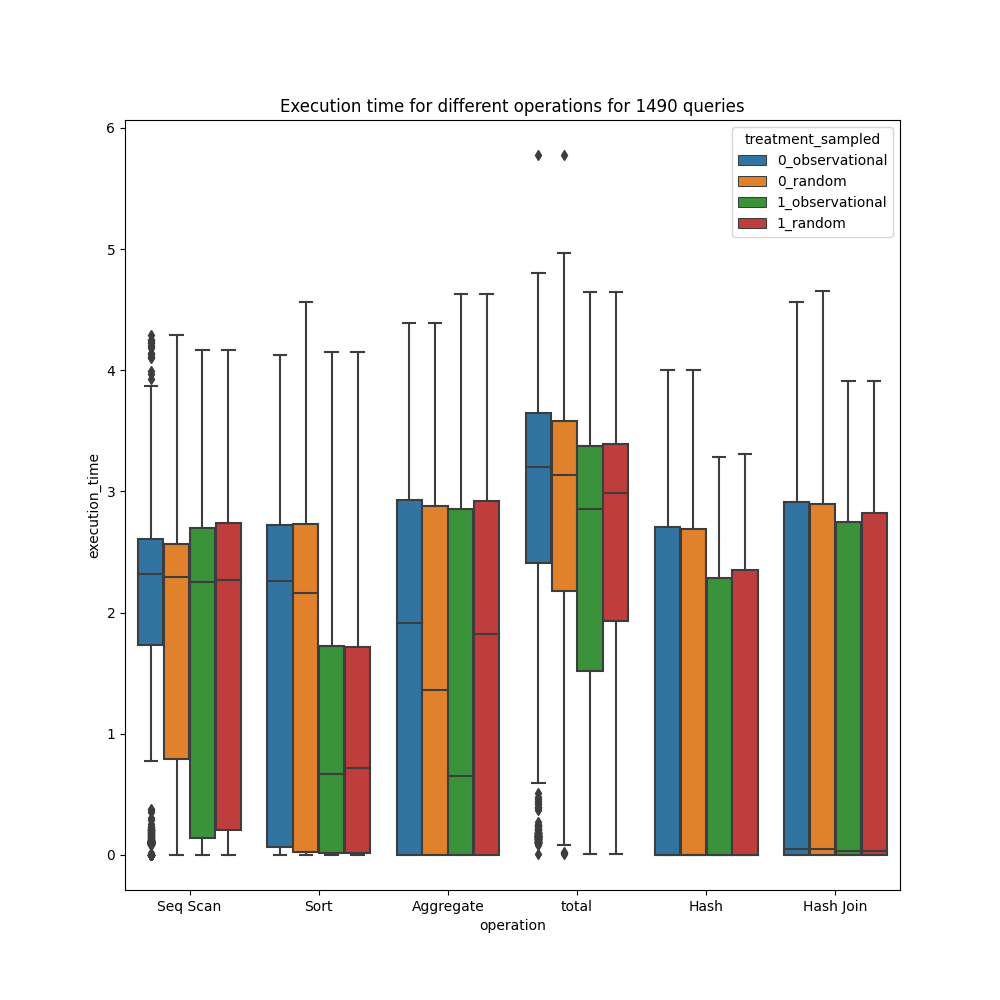}}
\caption{\textbf{Treatment effect on query executions for 1500 queries:} Ground-truth causal effect estimate of increasing memory for experimental data (random) and observational data created with bias strength = 1. 0 means low memory, and 1 means high memory. We can see that increasing memory has the most effect on sort and aggregate operation and the least effect on the sequential scan.}
\label{fig:ground_truth_effect_query_execution}

\vskip -0.2in
\end{figure}

We first collect $10000$ most popular user-defined \href{https://data.stackexchange.com/}{Math Stack Overflow} queries.  We install a \href{http://www.postgresql.org/}{PostgreSQL 14}  database server and load a 50 GB version of the publicly available \href{https://www.brentozar.com/archive/2015/10/how-to-download-the-stack-overflow-database-via-bittorrent/}{Stack Overflow Database}. We then run these queries with different combinations of the configuration parameters listed in Table \ref{tab:interventions}. In all our experiments, our queries were executed with PostgreSQL 14 database on a single node with an Intel 2.3 GHz 8-Core Intel Core i9 processor, 32GB of RAM, and a solid-state drive. PostgreSQL was configured to use a maximum of 0 parallel workers to ensure non-parallelized executions so that additive assumption about operations is satisfied (\texttt{max\_parallel\_workers\_per\_gather = {0}}). Before each run of the query, we begin from the cold cache by restarting the server to reduce caching effects among queries. Many database management systems provide information about the query plans as well as actual execution information through convenient APIs, such as \texttt{EXPLAIN ANALYZE} queries. Usually, Postgres reports the total run-time of each operation, along with children's operations. We mainly model the query plans with the following operations --- {Sequential Scan, Index Scan, Sort, Aggregate, Hash, Hash Join} as the occurrence of these operations in collected query plans was good, providing a large number of samples to learn the models from data. For CATE estimation experiments, we select $1500$ query plans in which effect sizes were significant and were actually a result of the intervention rather than random variation in the run-time due to the stochastic nature of the database execution system. Each SQL query is run $5$ times, and the median execution time is taken as the outcome. We evaluate the combined treatment effect of increasing working memory size and adding indices on structured query execution plans (Table \ref{tab:interventions}). Increasing working memory affects the run-time of sorting, hash join, and aggregation operations, as can be seen in Figure \ref{fig:ground_truth_effect_query_execution}. Adding additional indices modifies the structure of the execution plan by switching from a sequential scan component to an index scan component, as discussed below. 

\begin{table}[h]
    \centering
    \begin{tabular}{|c|c|c|c|}
    \midrule
    Treatment & Working Memory & Temp Buffers & Indices \\
    \midrule
          T=0 & 64 KB & 800 KB& Primary key indexing\\
          T=1 & 50 MB & 100 MB & Secondary key indexing\\
     \midrule
    \end{tabular}
    \caption{Treatment details for query execution data set}
    \label{tab:interventions}
\end{table}

\textbf{Change in the structure of query execution plans as a result of interventions on configuration parameters:} For some interventions on the configuration parameters and for some queries, the query planner doesn't return the same query plan. It returns the query plan with a changed structure as well as modified features of the components. This makes sense as that is the goal of query optimizers to compare different plans as resources change and find the most efficient plan. For example, increasing the working memory often causes query planners to change the ordering of Sort and aggregate operations, changing the structure as well as inputs to each component. These interventions are different from standard interventions in causal inference in which we assume that the covariates of the unit remain the same (as they are assumed to be pre-treatment) and treatment only modifies the outcome. In this case, a few features of the query plan are modified as a result of the intervention (and thus are post-treatment), while other features remain the same. Prediction of which features would change is part of learning the behavior of the query planner under interventions. In this work, we have mostly focused on learning the behavior of the query execution engine and assumed that the query planner is accessible to us. For simplicity, we assume that we know of the change in structure as a result of the intervention for both models. We leave the learning of the behavior of query optimizers under interventions for future work. This case provides another challenge for the task of causal effect estimation, even in the case of randomized treatments (bias strength = 0); due to the modified features of the query plans, the distribution of features in control and treatment populations might differ, providing an inherent observational bias in the dataset coming from the query optimizer. As long as we provide the information about modified query plans for both models, we believe that our comparisons are fair. For changed query structure, CATE estimand can be thought of as conditional on the same query but two different query plans. 
$$\tau(Q_{i}) = \mathbb{E}[Y_i(1) - Y_i(0) | Q_{i}]$$
$$\tau(Q_{i}) = \mathbb{E}[\mathbb{E}[Y_i(1) |Q_{p_i}(1)] - \mathbb{E}[Y_i(0) |Q_{p_i}(0)]]$$

\textbf{Covariates used for query execution data for model training:} See Table \ref{tab:input_features} for information about the high-dimensional covariates and component-specific covariates used for training models on query execution plans data set.
\begin{table}[ht]
     \centering
     \begin{tabular}{|p{0.15\linewidth}||p{0.1\linewidth} | p{0.5\linewidth}|p{0.18\linewidth}|}
     \midrule
     Model & Component& Training features  & Outcome \\
     \midrule
          Unitary &   & num\_Sort, num\_Hash\_Join, num\_Seq\_Scan, num\_Hash,
       num\_Index\_Scan, num\_Aggregate, num\_complex\_ops,
       Sort\_input\_rows, Hash Join\_input\_rows, Hash Join\_left\_plan\_rows,
       Hash Join\_right\_plan\_rows, Seq Scan\_input\_rows, Hash\_input\_rows,
       Index Scan\_input\_rows, Aggregate\_input\_rows & total\_time\\
           \midrule
           Compositional & Sequential Scan & Seq\_Scan\_input\_rows, Seq\_Scan\_plan\_rows& seq\_scan\_time\\
           \midrule
           Compositional & Index Scan & Index\_Scan\_input\_rows, Index\_Scan\_plan\_rows& index\_scan\_time\\
           \midrule
           Compositional & Hash & Hash\_input\_rows, Hash\_plan\_rows& hash\_time\\
            \midrule
           Compositional & Hash Join & Hash\_Join\_left\_input\_rows, Hash\_Join\_right\_input\_rows, Hash\_Join\_plan\_rows& hash\_join\_time\\
            \midrule
           Compositional & Sort & Sort\_input\_rows, Sort\_plan\_rows& sort\_time\\
            \midrule
           Compositional & Aggregate & Aggregate\_input\_rows, Aggregate\_plan\_rows& aggregate\_time\\
           \midrule
          
      \midrule
     \end{tabular}
     \caption{Covariates used by unitary and compositional models for query execution plans data set}
     \label{tab:input_features}
 \end{table}

\subsubsection{Manufacturing Plant Data}

\begin{figure}[h]
    \centering
    \includegraphics[width=\linewidth]{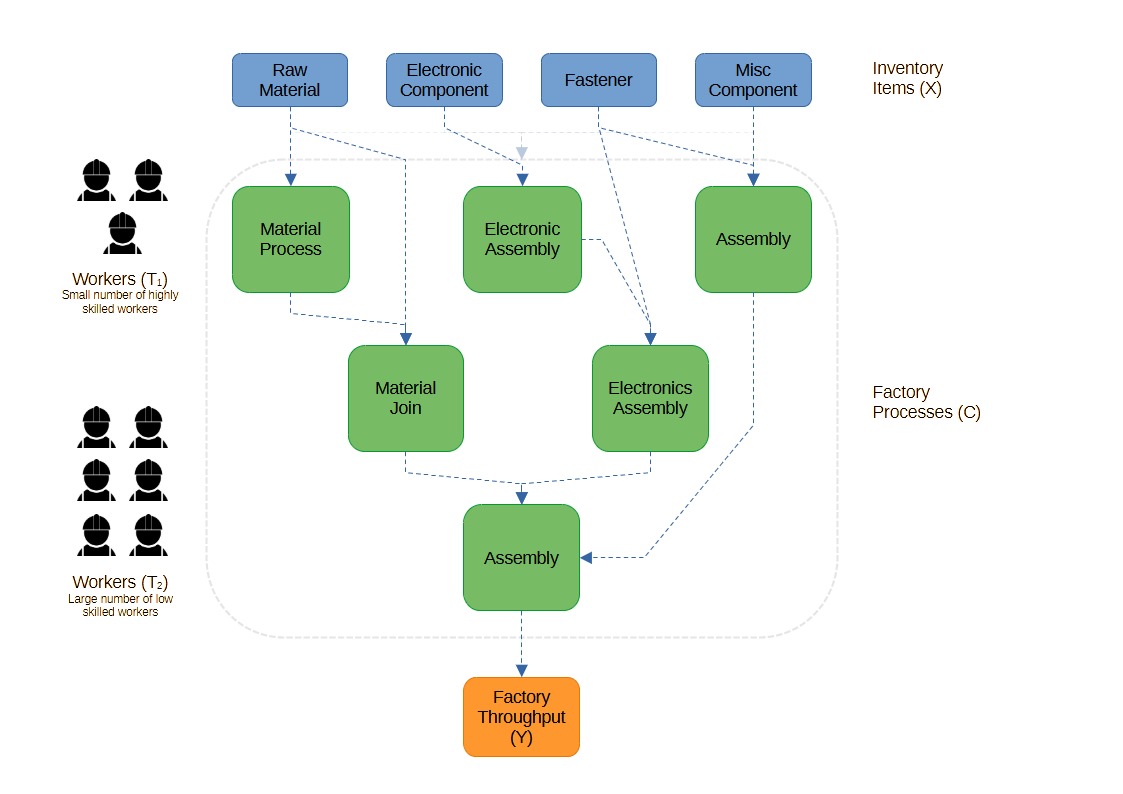}
    \caption{Illustrative figure explaining manufacturing assembly system}
    \label{fig:manufacturing_intro}
\end{figure}
\begin{figure}[h]
    \centering
    \includegraphics[width=\linewidth]{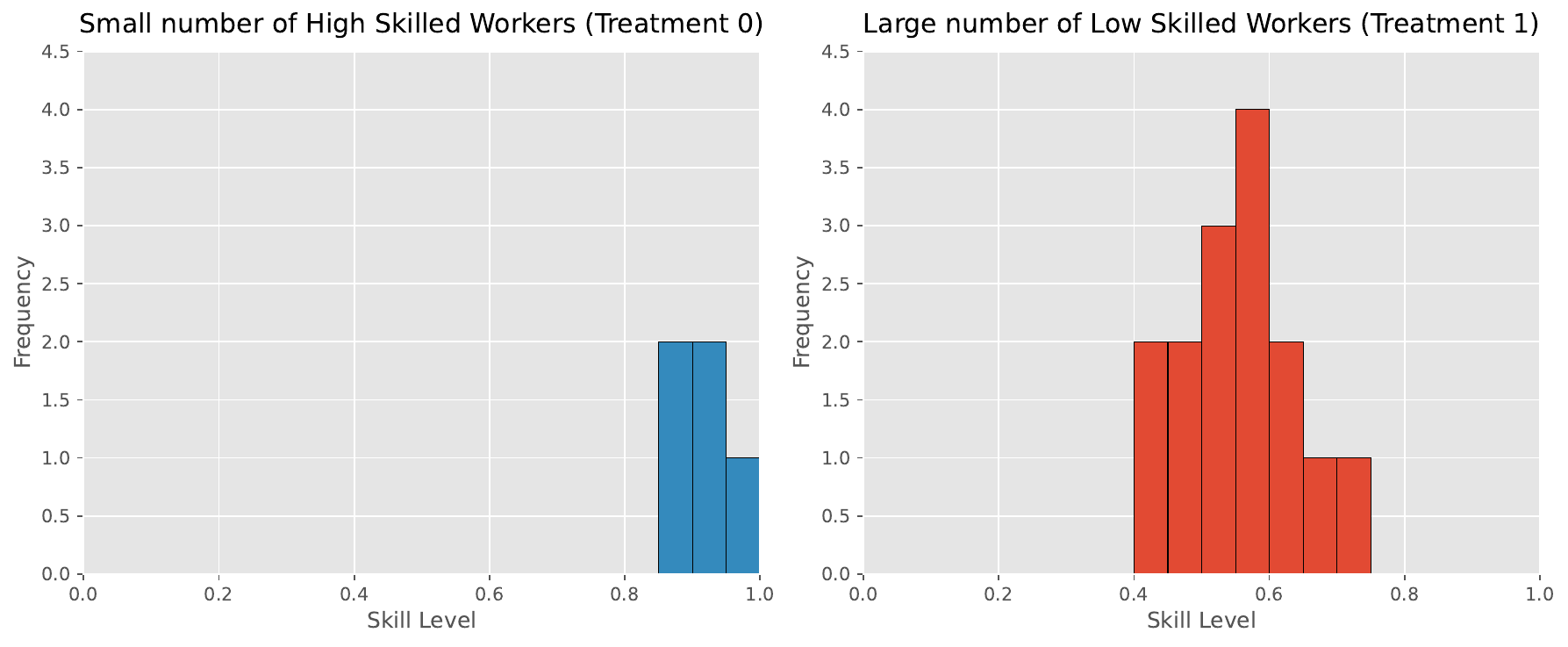}
    \caption{\textbf{Treatment for manufacturing plant data:} skill distribution for five (5) vs. fifteen (15) skilled workers.}
    \label{fig:worker_skill_disrtribution}
\end{figure}

\begin{figure}[h]
    \centering
    \includegraphics[width=\linewidth]{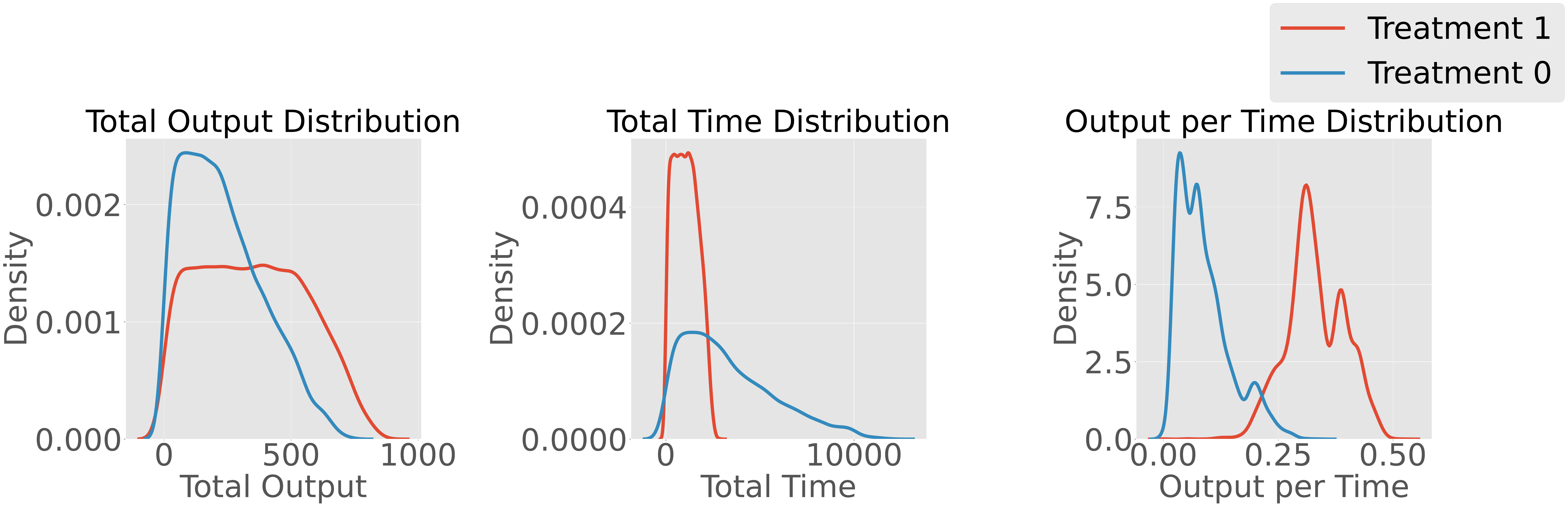}
    \caption{\textbf{Distribution of different potential outcomes for manufacturing data :} This figure shows the distribution of various potential outcomes for treatment and control groups for manufacturing plant data --- (a) total output (total number of parts produced), (b) total time (total processing time of each scenario), and (c) output per time (total parts produced/time) distributions. We consider total output (total number of parts produced (a)) in our experiments.}
    \label{fig:manufacturing_po}
\end{figure}
We use a process-based discrete-event simulation framework, \href{https://simpy.readthedocs.io/en/latest/}{Simpy}, to generate realistic manufacturing plant data. The plant aims to produce the final product by processing raw materials and assembling intermediate parts. The simulation comprises \textbf{\textit{four}} distinct manufacturing processes: Material Processing, Material Joining, Electronics Processing, and Assembly combined and re-used across $50$ manufacturing line layouts with varying hierarchical structures. Each scenario consists of simulations with product demand varying from $5$ to $1000$, with different raw material inventories (pre-treatment covariates) available for each demand. The intervention consists of the availability of multiple workers with two different skill levels -- (1) 5 workers with a higher mean of skill distribution and (2) 15 workers with a lower mean of skill distribution (Figure \ref{fig:worker_skill_disrtribution}). The goal is to estimate the effect of workers' (with different skill levels) availability on the number of parts produced. Figure \ref{fig:manufacturing_po} shows the effect of treatment on total output (total number of parts produced), total time (total processing time of each scenario), and output per time (total parts produced/time) distributions. This data satisfies causal Markov dependence among the potential outcomes as the quality of the part processed by the component is explained by the raw-material inputs, intervention, and the quality of the directly connected parent components. There are a total of $39$ unit-level covariates and around $10$ covariates per component. 

A factory has the following features:
\begin{enumerate}
    \item \textbf{Inventory of raw items} (Covariates): This includes parts and items which come into the factory to be processed into a final product. For this example, we use the following:
    \begin{enumerate}
        \item \textit{Fastener: }This includes parts such as screws, nuts and bolts.
        \item \textit{Electronic component:} electronic components are parts used in the assembly of electronic assemblies such as printed circuit boards.
        \item \textit{Raw material:} These are parts such as metal blanks, plastic sheets, sheet-metal etc.
        \item \textit{Misc Component:} Parts such as belts, pulleys, gears etc.
    \end{enumerate}

    \item \textbf{Process archetypes} (Components): These are processing stations which consume inventory of raw items and parts from other processes to produce a part. A process is defined by its processing time, i.e. how long does it ideally take to produce the part, and complexity, i.e how complex the given task is. Archetypes used in this simulation are:
    \begin{enumerate}
        \item \textit{Material Processing:} takes 1 part, does some processing and outputs the processed part (eg: painting, coating, heat treatment etc.)
    \item \textit{Material Joining:} takes two parts and joins them together to output another part (eg: welding, riveting, fasteners to put two parts together)
    \item \textit{Electronics Processing: }takes multiple components and electronics to produce a electronic part (eg: soldering, PCB manufacturing)
    \item \textit{Assembly:} takes components from other stations and puts them together to form an assembly (eg: final product assembly)
    \end{enumerate}

    \item \textbf{Workers} (Treatment): workers are a common resource pool of people working on a process. At a given time one process can have only one worker. A worker is defined by their skill which directly impacts how long a process will take over its base time and how much scrap (the final part is unusable and cannot proceed to the next step) and rework (redo aspects of the process increasing the amount of time the process takes) the process may produce.
\end{enumerate}

\subsubsection{Matrix operations processing}
\begin{figure}[h]
    \centering
    \includegraphics[width=\linewidth]{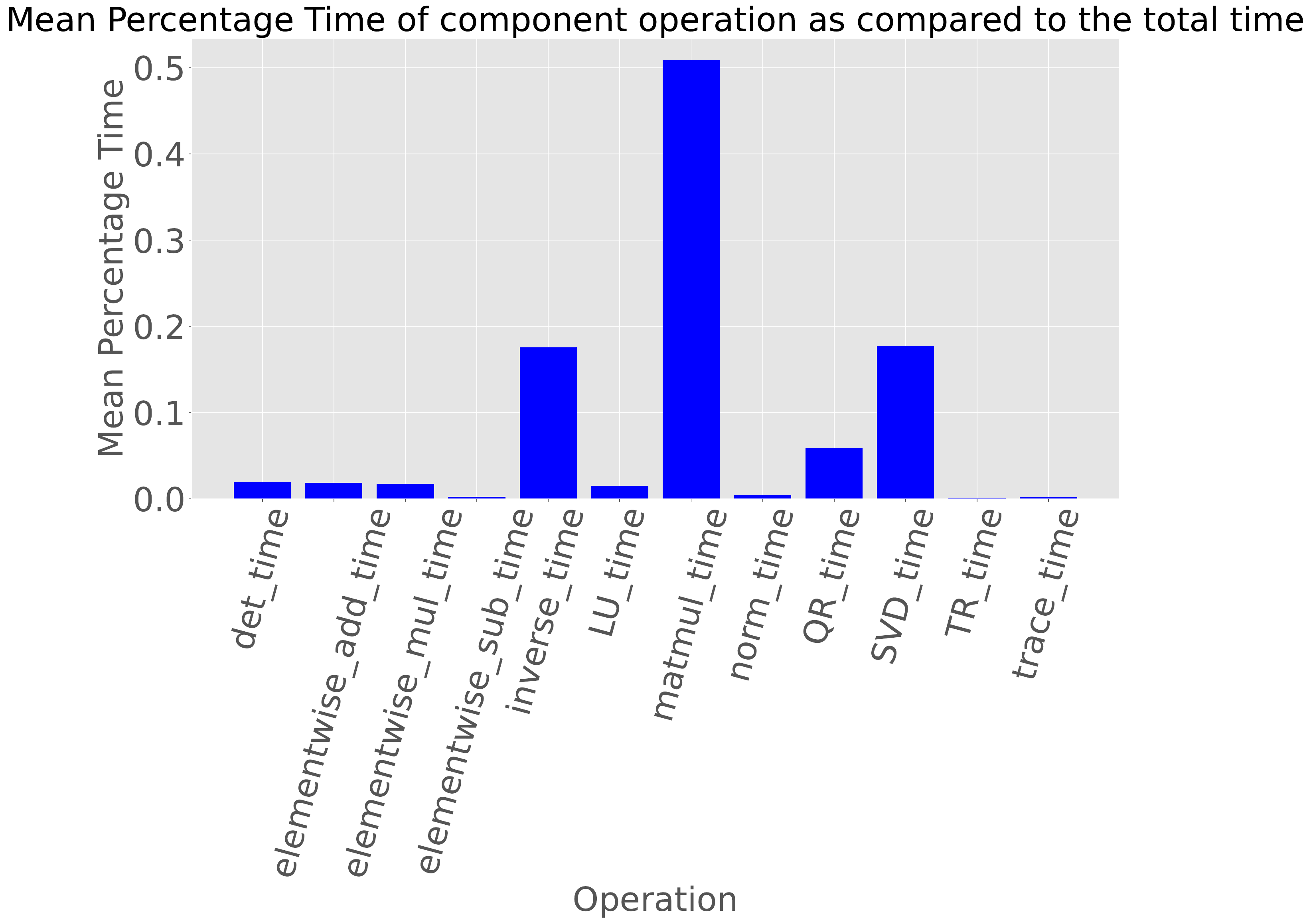}
    \caption{\textbf{Dominant contribution of component operation in total execution time:} This figure shows the dominant contribution of matrix multiplication's execution time in total execution time. Percentage contribution of each component is calculated for each unit. Mean is taken over 25000 unit instances corresponding to $25$ expression structrues. }
    \label{fig:dominant_matmul}
\end{figure}

\begin{figure}[h]
    \centering
    \includegraphics[width=\linewidth]{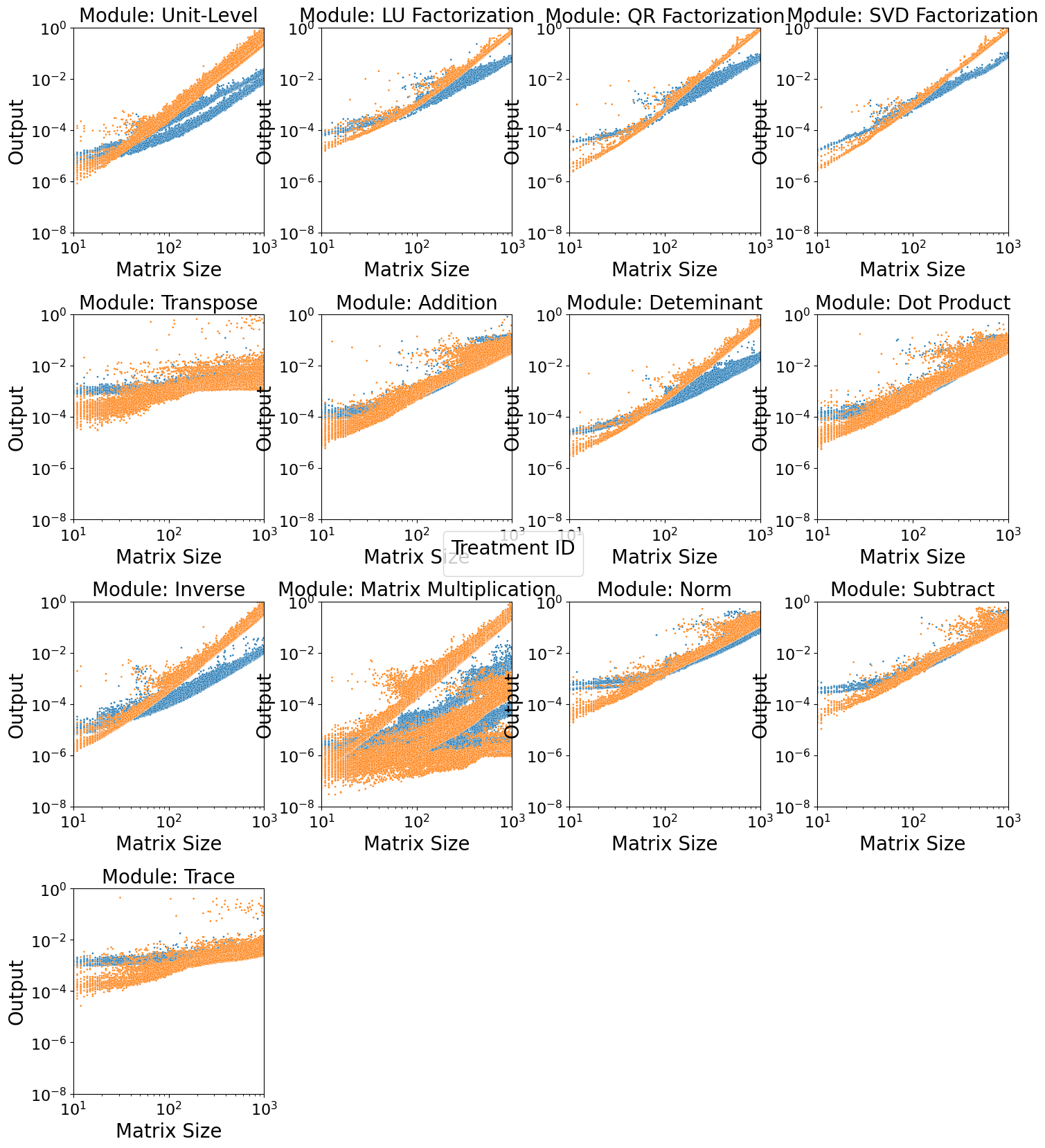}
    \caption{\textbf{Homogeneous functional forms of the component outcome function for matrix operations data set:} Ground-truth outcome functions for components for matrix operation data set. Blue and orange colors correspond to binary treatments $0$ and $1$. Intervention corresponds to two different compute hardware on which matrix expressions are processed.}
    \label{fig:matrix_mult_plots}
\end{figure}

We generate a matrix operations data set by evaluating complex $25$ complex matrix expressions on two computer hardware with different processors and RAM (treatment) and evaluate the execution time for each treatment (potential outcomes). The matrix size of matrices varies from $2$ to $1000$, resulting in total $25000$ unit samples. The expressions contain $12$ component operations, e.g., matrix multiplication, inverse, singular value decomposition, etc. We ensure each operation is executed individually, ensuring parallel composition with additive aggregation function. Matrix size is used as a biasing covariate to create a distribution mismatch between treatment groups. Treatment 0 means operations are processed on computer hardware with an 8-core, 32GB of RAM, and treatment $1$ means operations are processed on 1-core, 4 GB RAM.  Figure \ref{fig:matrix_mult_plots} shows the potential outcome functions for each treatment for unit-level and component data. Figure \ref{fig:dominant_matmul} shows that the matrix multiplication operation is the dominant operation, taking the most of the execution time (50\%) across all matrix expressions

\subsubsection{Synthetic data generation:}
We generate data sets with varying characteristics to test model performance for units with different structures and composition functions. Structured units are generated by sampling binary trees (max depth=$10$) with $k$=$10$ heterogeneous modules, each having $d_j$=$1$ feature ($d$=$10$ (covariates) + $10$ (structural information) total). All components' total sum of features is used as a biasing covariate to create a distribution mismatch based on joint covariates distribution. For observational bias based on the structure, the depth of the trees is used as a proxy for structural information. The covariate distribution for each component is sampled from a multivariate Gaussian and uniform distribution with a mean ranging between 0 and 3 and covariance ranging between 0 and 3. The potential outcome for each treatment is a polynomial smooth function with different parameters for each treatment to generate heterogeneous treatment effects. For fixed structure data generation, the depth of the tree is fixed to $10$ so that every unit has the same number and kind of components. For the variable structure setting, the depth of the tree randomly varies between $4$ and $10$, and components are sampled with replacement. For compositional generalization, an equal number of trees are generated for every module combination from $2$ to $10$. For parallel composition, the potential outcome is simulated for each component for each treatment as a function of input features and treatment. For sequential and hierarchical composition, the potential outcome is a function of input features, treatment, and potential outcomes of the parent components. 

\subsection{Models and Baseline implementation}
\begin{enumerate}
    \item \textit{Hierarchical composition models:} Each distinct component in a structured unit is implemented as a separate MLP module with a hidden dimension size = $2 * input\_size$ and batch size = $64$ for each component. Models were trained using Adam optimizer with a learning rate of $0.01$ (and adaptive cosine learning rate schedule with a starting learning rate of $0.001$ for certain data sets). The total mean squared loss for all the component outcomes was optimized for the observed fine-grained outcomes of the hierarchical model. In contrast, loss for only unit-level potential outcomes was optimized for the unobserved fine-grained outcomes model. For the unobserved fine-grained outcomes model, the outcomes are passed hierarchically through the component interaction graph $G_i$ as illustrated in Figure \ref{fig:intro_graphics}(c). 
    \item \textit{Additive parallel composition models:} We implement an additive parallel model using two model classes: random\_forest and neural\_network. The differences between hierarchical and additive parallel composition models are as follows: (1) In parallel composition, the outcomes of each component are computed independently and not shared hierarchically. Second, an additive aggregation function is assumed, i.e., the unit-level outcome is the sum of the component's outcomes. A three-layer, fully connected MLP architecture is used for neural network models with hidden layer dimension = 64 and ReLU activations. Models were trained using Adam optimizer with a learning rate of $0.01$. For unobserved component-level outcomes, a mixture of experts (MoE) architecture \citep{jacobs1991adaptive, shazeer2016outrageously} is used where each expert receives the high-dimensional covariates $\mathbf{X} \in \mathbb{R}^d$, initialized with the same number of experts as a number of distinct modules in the domain. We use a simple addition of the experts' outcomes for the gating mechanism, as our data sets satisfy additive composition. For unobserved component-level covariates, entire unit-level covariates are passed to each component model to learn component-specific representations as part of learning component outcome functions. 
\end{enumerate}

\textbf{Baselines:} X-learner and non-parametric double machine learning implementation is from \href{https://github.com/py-why/EconML}{Econml} library and random forests were used as the base models. 
TNet \citep{curth2021nonparametric} implementation is taken from the GitHub repository \href{https://github.com/AliciaCurth/CATENets}{catenets}. 

\section{Additional Results}
\begin{figure}[h]
    \centering
    \includegraphics[width=\linewidth]{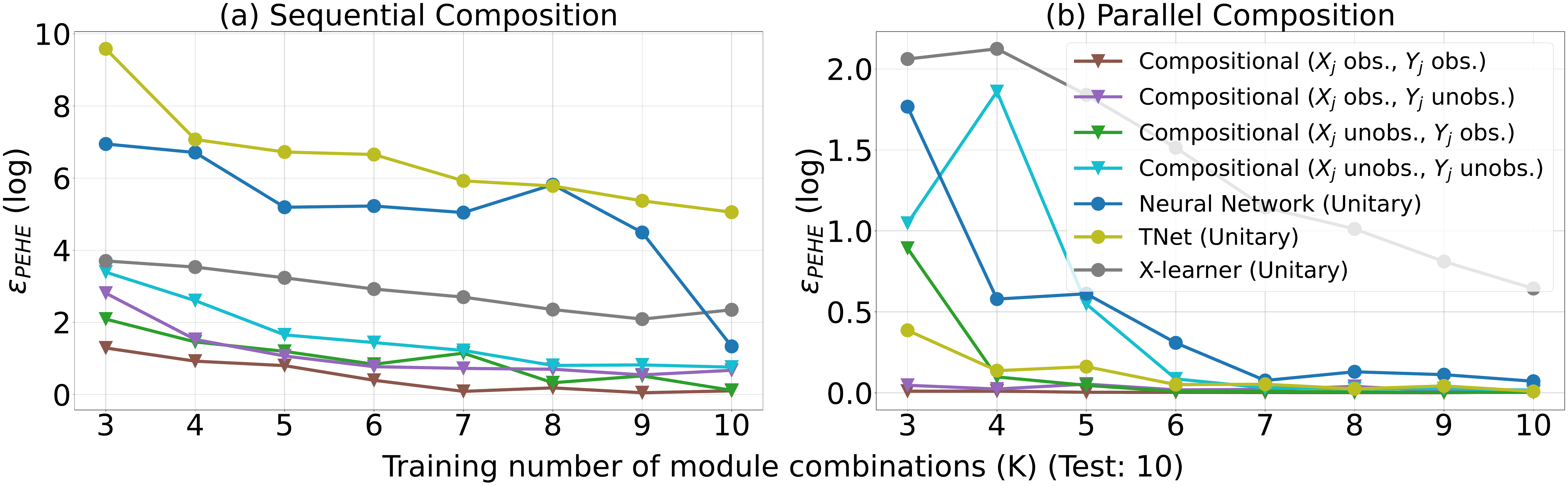}
    \caption{\textbf{Role of component-level data access and composition structure in the performance of
compositional models:}PEHE error (in log) for models evaluated on compositional generalization task with varying degrees of component-level data access. (Lower is better). We observe that end-to-end trained models incorporating just modular structure compositionally generalize as trained on more module combinations. Unitary models show compositional generalization for additive parallel composition but perform comparably only for in-distribution combinations ($K$=$10$) for sequential composition, except X-learner. Note that the number of training samples increases as training depth increases.}
    \label{fig:prior_knowledge_pehe_app}
\end{figure}


\begin{figure}
    \centering
    \includegraphics[width=\linewidth]{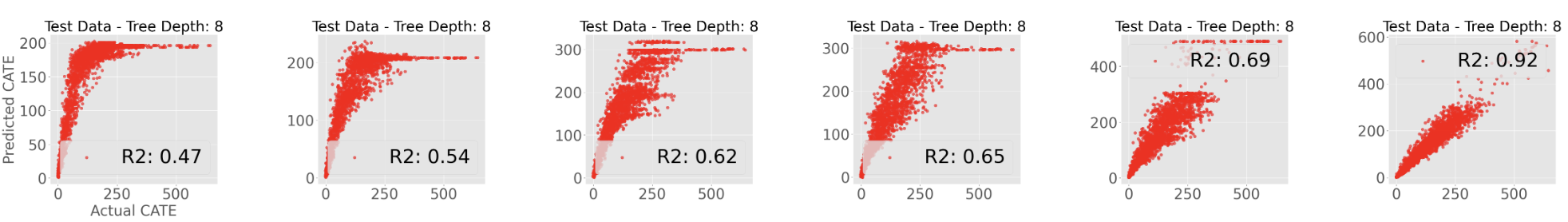}
    \caption{\textbf{Compositional generalization scatter plot for unitary model (X-learner):} This figure shows the scatter plot for test performance for the compositional generalization experiment. The training depths are varied from left to right $K$ = $3$ to $8$ (left to right)}
    \label{fig:manufacturing_po_scatter_high_level}
\end{figure}

\begin{figure}[h]
    \centering
    \includegraphics[width=\linewidth]{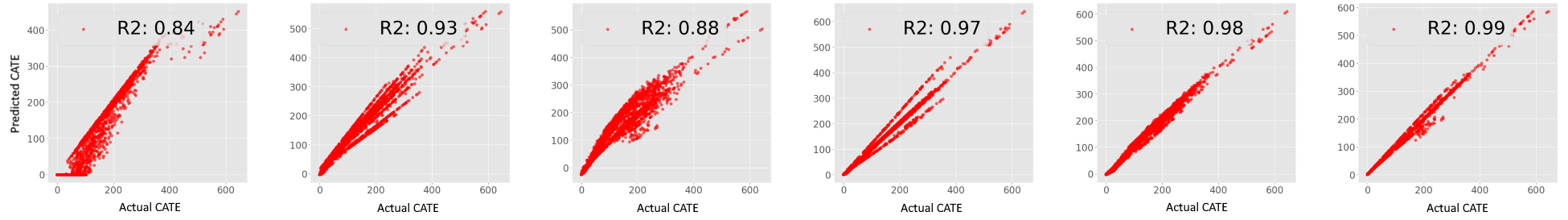}
    \caption{\textbf{Compositional generalization scatter plot for compositional model:} This figure shows the scatter plot for the predicted \textbf{test} performance of the compositional model on the test depth=8 when trained on increasing depths $K$ = $3$ to $8$ (left to right).}
    \label{fig:manufacturing_po_scatter_low_level}
\end{figure}

\begin{figure}[h]
    \centering
    \includegraphics[width=\linewidth]{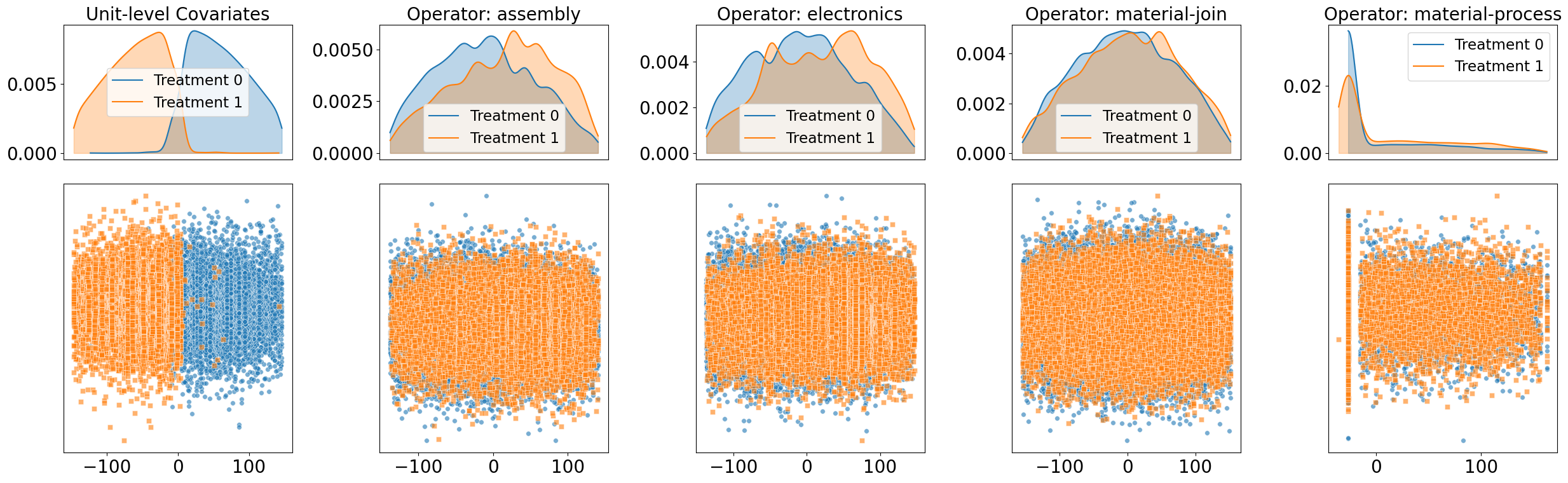}
    \caption{\textbf{Distribution mismatch due to structure-based treatment assignment for manufacturing data}. These density plots (top) and scatter plots (bottom) show the distribution mismatch between treatment and control groups at the unit level and component level. The plots are generated by first projecting the high-dimensional unit-level and component-level covariates to 1-dimensional using TSNE. The observational bias is created using tree\_depth feature of the hierarchical structure.}
    \label{fig:dist_mismatch}
\end{figure}

\section{Algorithms}

\subsubsection{Hierarchical Composition Model}
\begin{algorithm}[ht]
\caption{Hierarchical Composition Model: Training}
\label{alg:ite_estimation_hierarchical_training_app}
\begin{algorithmic}[1]
\STATE {\bfseries Input:} Factual data set: $\mathcal{D}_F = \{q_i : \{\mathbf{x}_{ij}\}_{j=1:m_i}, t_i, y_i, \{y_{ij}\}_{j=1:m_i}\}_{i=1:n}$, number of distinct components $k$.
\STATE{\bfseries Result:} Learned potential outcome models for each component: $\{\hat{f}_{\theta_1}, \hat{f}_{\theta_2}, \hat{f}_{\theta_3} \dots \hat{f}_{\theta_k}\}$
 \WHILE{not converged}
    \STATE $loss\_1, loss\_2, loss\_3, loss\_k = 0$
    \FOR{$i=1$ {\bfseries to} ${n}$}
        \STATE Get the order of the components in which input is processed by using post-order traversal of the tree $G_i$.
        \STATE $orderedList \gets post\_order\_traversal(G_i)$  
        \FOR{component $j$ in $orderedList$}
        \STATE $o \gets component\_class\_index(j)$
        \STATE // Potential outcome of a component depends on the potential outcome of the parent components according to graph $G_i$ (assuming binary tree)
        \IF{component $j$ has parents in $G_i$}
        \STATE $\hat{y}_{ij} = \hat{f}_{\theta_o}(\mathbf{x}_{ij}, \{y_{lt_i}\}_{l \in Pa(c_j)}, t_i)$
        \ELSIF{component $l$ does not have parents}
        \STATE $\hat{y}_{ij} = \hat{f}_{\theta_o}(\mathbf{x}_{ij}, t_i)$
        \ENDIF
        \STATE $loss\_o = loss\_o + (\hat{y}_{ij} - y_{ij})^2$
        \ENDFOR
        
    \ENDFOR
    \STATE Calculate gradients for the parameters for each module

    \FOR{$o=1$ {\bfseries to} ${k}$}  
    \STATE $\delta_o \gets \triangle_{\theta_o} \frac{1}{N_o} loss\_o$\;
    \STATE $\theta_o \gets \theta_o - \alpha \delta_o$\ independent training of all the component models.
    \ENDFOR
    \STATE Check convergence criterion
\ENDWHILE
\end{algorithmic}
\end{algorithm}

\begin{algorithm}[ht]
\caption{Hierarchical Composition Model: Inference}
\label{alg:ite_estimation_hierarchical_inference_app}

\begin{algorithmic}[1]
\STATE {\bfseries Input:} Test data set: $\mathcal{D_T} = \{q_i : \{\mathbf{x}_{ij}\}_{j=1:m_i}\}_{i=1:n}$, learned potential outcome models for each component: $\{\hat{f}_{\theta_1}, \hat{f}_{\theta_2}, \hat{f}_{\theta_3} \dots \hat{f}_{\theta_k}\}$,
\STATE{\bfseries Result:} CATESamples
\vspace{5pt}
\STATE {\bfseries Procedure:}
\STATE $CATESamples \gets \{\}$
\FOR{$i=1$ {\bfseries to} ${n}$}
    \STATE Get the order of the operation in which input is processed by post-order traversal of the tree
        \STATE $orderedList \gets post\_order\_traversal(G_i)$  
        
        \FOR{component $j$ in $orderedList$}
        \STATE $o \gets component\_class\_index(j)$
        \IF{component $j$ has parents in $G_i$}
        \STATE $\hat{y}_{ijt_i} = \hat{f}_{\theta^*_o}(\mathbf{x}_{ij}, \{\hat{y}_{ilt_i}\}_{l \in Pa(c_j)}, t_i)$
        \ELSIF{component $l$ does not have parents}
        \STATE $\hat{y}_{ijt_i} = \hat{f}_{\theta^*_o}(\mathbf{x}_{ij}, t_i)$
         \ENDIF
        \ENDFOR
    \STATE $\hat{\tau}(q_i) = \hat{y}_{im_i}(1) - \hat{y}_{im_i}(0)$, get the difference between estimated potential outcomes of the last component in $G_i$
    \STATE $CATESamples \gets CATESamples \cup \{(q_i, \hat{\tau}(q_i))\}$
\ENDFOR
\end{algorithmic}
\end{algorithm}

\begin{algorithm}[ht]
\caption{Additive Parallel Composition Model: Training}
\label{alg:ite_estimation_parallel_training}
\begin{algorithmic}[1]
\STATE {\bfseries Input:} Factual data set: $\mathcal{D}_F = \{q_i : \{\mathbf{x}_{ij}\}_{j=1:m_i}, t_i, y_i, \{y_{ij}\}_{j=1:m_i}\}_{i=1:n}$, number of distinct components $k$.
\STATE{\bfseries Result:} Learned potential outcome models for each component: $\{\hat{f}_{\theta_1}, \hat{f}_{\theta_2}, \hat{f}_{\theta_3} \dots \hat{f}_{\theta_k}\}$
\vspace{5pt}
\STATE {\bfseries Procedure:}

    \STATE $\mathcal{D}_1 \gets \{\}, \mathcal{D}_2 \gets \{\}, \mathcal{D}_3 \gets \{\} \dots \mathcal{D}_k \gets \{\} $
    \FOR{$i=1$ {\bfseries to} ${n}$}  
        \FOR{$j=1$ {\bfseries to} ${m_i}$} 
          \STATE $o \gets component\_class\_index(j)$ \; index of distinct component class for $j^{th}$ component instance.
          \STATE $\mathcal{D}_o \gets \mathcal{D}_o \cup \{\mathbf{x}_{ij}, t_i, y_{ij}\}$
        \ENDFOR
    \ENDFOR
    \FOR{$o=1$ {\bfseries to} ${k}$}  
    \STATE $N_o \gets len(\mathcal{D}_o)$\;
    \STATE $\theta_o := \arg\min_{\theta} \frac{1} {N_o}\sum_{m=1}^{N_o} (\hat{f}_{\theta_o}(\mathbf{x}_{m}, t_m) - y_{m})^2$ \;  independent training of all the component models.
    \ENDFOR
\end{algorithmic}
\end{algorithm}

\begin{algorithm}[ht]
\caption{Additive Parallel Composition Model: Inference}
\label{alg:ite_estimation_parallel_inference}
\begin{algorithmic}[1]
\STATE {\bfseries Input:} Test data set: $\mathcal{D_T} = \{q_i: \{\mathbf{x}_{ij}\}_{j=1:m_i}\}_{i=1:n}$ and potential outcome models for each component: $\{\hat{f}_{\theta_1}, \hat{f}_{\theta_2}, \hat{f}_{\theta_3} \dots \hat{f}_{\theta_k}\}$,
\STATE{\bfseries Result:} CATESamples
\vspace{5pt}
\STATE {\bfseries Procedure:}
\STATE $CATESamples \gets \{\}$
\FOR{$i=1$ {\bfseries to} ${n}$}
    \FOR{$j=1$ {\bfseries to} ${m_i}$}
          \STATE $o \gets component\_class\_index(j)$
          \STATE $\hat{y}_{ij1} \gets \hat{f}_{\theta^{*}_o}(\mathbf{x}_{ij}, 1)$
          \STATE $\hat{y}_{ij0} \gets \hat{f}_{\theta^{*}_o}(\mathbf{x}_{ij}, 0)$
    \ENDFOR
    \STATE $\hat{y}_{i1} = \sum_{j=1}^{m_i} \hat{y}_{ij1}$
    \STATE $\hat{y}_{i0} = \sum_{j=1}^{m_i} \hat{y}_{ij0}$
    \STATE $\hat{\tau}(q_i) = \hat{y}_{i1} - \hat{y}_{i0}$
    \STATE $CATESamples \gets CATESamples \cup \{(q_i, \hat{\tau}(q_i))\}$
\ENDFOR
\end{algorithmic}
\end{algorithm}

\end{document}